\documentclass[lettersize,journal]{IEEEtran}
\usepackage{amsmath,amsfonts}
\usepackage{algorithmic}
\usepackage{algorithm}
\usepackage{array}
\usepackage[caption=false,font=normalsize,labelfont=sf,textfont=sf]{subfig}
\usepackage{textcomp}
\usepackage{stfloats}
\usepackage{url}
\usepackage{verbatim}
\usepackage{graphicx}
\usepackage{cite}
\usepackage{xcolor}
\usepackage{multirow}
\usepackage{algorithm}
\usepackage{algorithmic}
\usepackage{amssymb}
\begin{document}

\title{Self-Paced Learning for Open-Set Domain Adaptation}


\author{Xinghong Liu, Yi Zhou,~\IEEEmembership{Member,~IEEE,} Tao Zhou, Jie Qin, and Shengcai Liao,~\IEEEmembership{Senior Member,~IEEE}
\thanks{Corresponding author: Yi Zhou}
\thanks{Xinghong Liu and Yi Zhou are with the School of Computer Science and Engineering, Southeast University, Nanjing, China (e-mail: xhoml158@gmail.com, yizhou.szcn@gmail.com).}
\thanks{Tao Zhou is with the School of Computer Science and Engineering, Nanjing University of Science and Technology, Nanjing, China (e-mail: taozhou.ai@gmail.com)}
\thanks{Jie Qin is with the College of Computer Science and Technology, Nanjing University of Aeronautics and Astronautics, Nanjing, China (e-mail: qinjiebuaa@gmail.com)}
\thanks{Shengcai Liao is with Inception Institute of Artificial Intelligence (IIAI), Abu Dhabi, U.A.E. (e-mail: shengcai.liao@inceptioniai.org)}
}

\markboth{Journal of \LaTeX\ Class Files}%
{Shell \MakeLowercase{\textit{et al.}}: A Sample Article Using IEEEtran.cls for IEEE Journals}


\maketitle

\begin{abstract}
Domain adaptation tackles the challenge of generalizing knowledge acquired from a source domain to a target domain with different data distributions. Traditional domain adaptation methods presume that the classes in the source and target domains are identical, which is not always the case in real-world scenarios. Open-set domain adaptation (OSDA) addresses this limitation by allowing previously unseen classes in the target domain.
Open-set domain adaptation aims to not only recognize target samples belonging to common classes shared by source and target domains but also perceive unknown class samples. We propose a novel framework based on self-paced learning to distinguish common and unknown class samples precisely, referred to as SPLOS (self-paced learning for open-set). To utilize unlabeled target samples for self-paced learning, we generate pseudo labels and design a cross-domain mixup method tailored for OSDA scenarios. This strategy minimizes the noise from pseudo labels and ensures our model progressively learns common class features of the target domain, beginning with simpler examples and advancing to more complex ones. Furthermore, unlike existing OSDA methods that require manual hyperparameter $threshold$ tuning to separate common and unknown classes, our approach self-tunes a suitable threshold, eliminating the need for empirical tuning during testing. Comprehensive experiments illustrate that our method consistently achieves superior performance on different benchmarks compared with various state-of-the-art methods.
\end{abstract}


\section{Introduction}
The most attractive point of deep learning methods is that models \cite{ResNet, VGG, Faster_RCNN, Mask_RCNN} trained on millions of annotated samples can achieve impressive performance on in-distribution new data. However, the performance of trained models usually dramatically drops when they are deployed on the target domain whose domain distribution is significantly different from the source domain. The mismatched domain distribution is called domain shift. It is caused by various perspectives, illumination conditions, and sensors, but can be minimized by transfer learning methods \cite{Dataset_Shift_in_Machine_Learning, A_Survey_on_Transfer_Learning} exploiting the information of labeled target samples. An intuitive way is to label target samples and fine-tune the trained model.

Nevertheless, annotating massive samples on the target domain is time-consuming and costly. To tackle this issue, unsupervised domain adaptation (UDA) approaches \cite{DAN,DANN,DDAN,DAAN} have been proposed to transfer knowledge from an annotated domain to an unlabeled domain. A classic scheme \cite{DANN,UAN,CMU} of UDA utilizes domain-hard labels (i.e. marking source samples as 1 and target samples as 0) and then applies a gradient reverse layer to minimize the marginal distribution divergence between source and target domains. In practice, partial samples in the target domain may not belong to any category in the source domain. Thus, more recently, open-set domain adaptation (OSDA) proposed in \cite{ATI,OSBP} requires to differentiate those unknown samples whose classes do not appear in the source domain, since forcing to adapt unknown classes with source domains will inevitably lead to negative transfer in OSDA scenarios. We observe that the domain-hard labels will erode the performance of the models. Roughly adopting the domain-hard labels for adaptation will match the whole target domain with the source domain, so the target-private class samples will be wrongly predicted as common classes shared by the source and target domains. 
To tackle the extra challenge of OSDA, we introduce a novel framework based on self-paced learning to align common classes in the target domain with the source domain, learn domain-invariant features, and differentiate between common and unknown classes in the target domain.

To align common classes in the target domain with the source domain, we propose a module consisting of dual multi-class classifiers (DMC). On the one hand, DMC can guide the model to only align common classes of the target domain instead of the whole target domain with the source domain, preventing negative transfer associated with aligning the entire target domain to the source domain. On the other hand, utilizing unlabeled target samples, DMC can gradually tune the threshold based on the confidence toward common classes in different training phases.
To learn the common class features of the target domain, based on self-paced learning, we introduce cross-domain mixup with multiple criteria (CMMC) for minimizing the noise caused by pseudo labels and learning domain-invariant features. CMMC utilizes a cross-domain mixup between source samples with ground truth labels and target samples with pseudo-labels to build a cross-domain bridge for OSDA tasks. During the initial training phase, DMC calculates a high threshold, resulting in low noise from the pseudo-labels. As training progresses, the threshold is gradually adjusted to a suitable value, allowing the model to obtain more diverse target domain samples with pseudo-labels. Therefore, the model can learn from the target samples of common classes, progressing from easier to more complex examples. While the noise from pseudo-labels may increase, the learning rate is already relatively low, mitigating the negative impact on model performance. To differentiate between common and unknown classes in the target domain, we combine various criteria, including entropy, consistency, and confidence. Relying solely on confidence values is inadequate for accurately distinguishing common and unknown classes, as confidence lacks discriminability for degrees of uncertainty. Therefore, we combine entropy and confidence, which are complementary and cover both smooth and non-smooth class distributions, and consistency, which can compensate for the confidence in prediction errors \cite{CMU}. In summary, we propose the SPLOS framework based on self-paced learning to align the common classes of the target domain with the source domain, learn domain-invariant features, and distinguish common and unknown classes. Our contributions are summarized as follows:

1. To effectively screen samples from simple to complex and avoid negative transfer caused by domain-hard labels, we introduce dual multi-class classifiers (DMC). Instead of adapting the entire target domain to the source domain, the DMC focuses on matching common classes between the two domains. Additionally, the DMC automatically computes an instructive threshold for discriminating common and unknown classes in the target domain. This self-tuning threshold enables the model to learn from target samples of common classes, progressing from easier to more complex examples. Our method eliminates the need for empirical tuning of the optimal threshold during the testing phase to distinguish common and unknown samples.

2. We introduce a novel approach, cross-Domain mixup with multiple criteria (CMMC), based on self-paced learning, which effectively learns domain-invariant features and distinguishes between common and unknown classes. By leveraging cross-domain mixup, CMMC significantly enhances our model's ability to learn domain-invariant features, while minimizing the noise caused by pseudo labels. Utilizing diverse criteria proves advantageous in distinguishing common and unknown samples compared to relying solely on confidence values.

3. We conducted comprehensive experiments to compare our method with various state-of-the-art techniques on three public benchmark datasets. Our model consistently achieves superior performance, illustrating its effectiveness. Furthermore, we thoroughly analyze the impact of each proposed component in our method to better understand their contributions to the overall performance.

\section{Related Work}
\subsection{Unsupervised Domain Adaptation}
Unsupervised Domain Adaptation (UDA) performs model training on the target domain without any label information to alleviate performance degradation caused by domain discrepancy. The mathematical essence of UDA is to minimize the joint distribution shift, which can be divided into marginal distribution shift and conditional distribution shift. Ben-David $et\ al.$ \cite{Analysis_of_Representations_for_Domain_Adaptation} theoretically proved that the goal of UDA can be achieved by reducing the inter-domain divergence while maximizing the margin of different categories on the source domain at the same time. Inspired by the generative adversarial network \cite{GAN}, Ganin $et\ al.$ \cite{DANN} designed an adversarial domain module to measure domain divergence and introduced a gradient reverse layer (GRL) to minimize the marginal distribution shift between source and target domains. GRL can help the model learn domain-invariant features. Wu $et\ al.$ \cite{DMRL} clarified that the limited number of samples from source and target domains could not guarantee that features in the latent spaces are domain-invariant. They introduced a cross-domain and inter-category mixup method to guide the classifier in learning domain-invariant features in a more continuous latent space. Xu $et\ al.$ \cite{DM-ADA} proposed a cross-domain mixup method on pixel and feature levels. They applied the mixup method in source and target samples with different ratios to generate various features representing different states between domains. \cite{DMRL, DM-ADA} generate a more continuous latent space to guarantee that features are domain-invariant for closed-set domain adaptation after minimizing the marginal distribution shift. Long $et\ al.$ \cite{CDANs} considered that reducing marginal distribution divergence probably cannot precisely align two domains with the multi-modal distribution. They constructed a framework with a conditional discriminator to reduce the conditional distribution difference. Yu $et\ al.$ \cite{DAAN} clarified that the importance of marginal and conditional distributions in real applications is different. They proposed a dynamic adversarial factor to quantitatively evaluate the relative importance of the marginal and conditional distributions. They improved their model performance compared with DANN \cite{DANN}. These traditional UDA methods are designed for the closed-set domain adaptation (CSDA) task, which cannot be directly applied to open-set domain adaptation problems.

\subsection{Open-Set Domain Adaptation}
Compared with CSDA tasks, the additional challenge for OSDA is that models need to split the target-private classes from the common classes without annotation in target domain. A representative CSDA method labels source and target samples as 1 and 0, respectively, and design a GRL to reduce the marginal distribution divergence of the two domains. Nevertheless, it is not sensible to match the whole target domain with the source domain in OSDA scenarios since it will cause the model to classify the unknown samples into common classes. Saito $et\ al.$ \cite{OSBP} designed a classifier with an additional class $unknown$ to discriminate the categories which only exist in the target domain. They proposed an optimization objective with an empirical hyperparameter to train the classifier. Liu $et\ al.$ \cite{STA} adopted a coarse-to-fine weighting mechanism to gradually split common and unknown classes in the target domain. Their approach allows weighing the importance of different samples while employing domain adaptation. They also introduced openness, which measures the proportion of unknown classes in all target classes. Shermin $et\ al.$ \cite{DAMC} also had a similar idea like \cite{STA}. A supplemental classifier is introduced to assign different weights to each sample. Luo $et\ al.$ \cite{PGL} proposed a novel method using a graph neural network with episodic training to repress underlying conditional shifts. Subsequently, they adopted adversarial learning to minimize the divergence between source and target domains. Wang $et\ al.$ \cite{SE-CC} designed a novel framework called self-ensembling with category-agnostic clusters (SE-CC). They clustered all unlabeled target samples to acquire category-agnostic clusters, which assist in disclosing the underlying feature space structure associated with the target domain. Moreover, they apply mutual information to improve the model performance. All the above OSDA methods distinguish common and unknown classes \textcolor{black}{relying on the confidence produced by the model in the testing phase}. It is not intensely reliable in some scenarios, especially when the openness of the target domain is large. By contrast, we propose to learn domain-invariant features in a more continuous latent space and combines multiple criteria to precisely separate common/unknown classes.

\subsection{Self-Paced Learning}
The self-paced learning paradigm, centered around an "easy-to-complex" training approach, lies at the heart of numerous supervised learning models, particularly those with noisy labels \cite{self-paced_learning_for_latent_variable_models} \cite{SFRS,CurriculumNet,Mentornet,ASPL}. Recently, efforts have been made to integrate self-paced learning with unsupervised domain adaptation, as seen in studies like PCDA \cite{PCDA} and SPCL \cite{ SPCL}. However, PCDA \cite{PCDA} is specifically designed for closed-set domain adaptation (CSDA), rendering it unsuitable for open-set domain adaptation (OSDA) scenarios. On the other hand, SPCL \cite{SPCL} assumes that target samples belong exclusively to unknown classes, implying no overlap between the target and source domain classes. This assumption also prevents the application of SPCL in OSDA situations where the target and source domains share common classes.

\section{Methodology}

\subsection{Preliminary}
In a standard unsupervised domain adaptation (UDA) scenario, we have a source domain $D^S=\{(x_i^s, y_i^s)\}^{N^S}_{i=1}$ of $N^S$ labeled samples and a target domain $D^T=\{(x_i^t)\}^{N^T}_{i=1}$ of $N^T$ unlabeled samples. The source domain distribution $p$ is different from the target domain distribution $q$. We define a labeled source class set $C^S$ and an unlabeled target class set $C^T$, which are subject to $C^S  \subsetneq C^T$ in open-set domain adaptation (OSDA) scenarios. $C=C^S \cap C^T$ is the common class set, and $C=C^S$ in OSDA scenarios. $C^{T\backslash S}=C^T \backslash C^S$ denotes the unknown class set. The openness is defined as $O=1-\frac{|C^S|}{|C^T|}$, where $|\cdot|$ is the cardinality of a set.

\subsection{Overview}
\begin{figure*}
	\centering
		\includegraphics[width=1.0\textwidth]{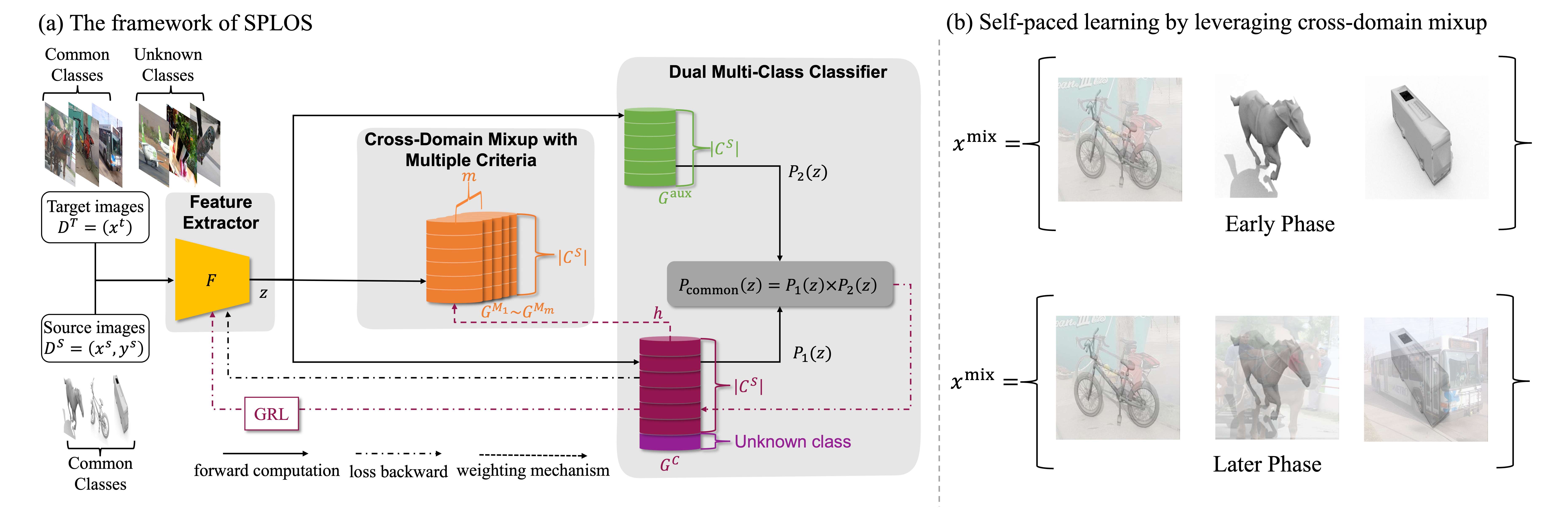}
	\caption{(a) The architecture of the \textcolor{black}{SPLOS (self-paced learning for open-set)}. $F$ is a backbone to extract features. GRL is a gradient reverse layer proposed in \cite{DANN} to adapt the target domain with the source domain. \textcolor{black}{SPLOS} contains two modules: \textcolor{black}{cross-domain mixup with multiple criteia (CMMC) module and dual multi-class classifier (DMC) module}. CMMC consists of $m$ classifiers from $G^{M_1}$ to $G^{M_m}$. \textcolor{black}{DMC} consists of two classifiers: an adversarial learning classifier $G^C$ and an auxiliary classifier $G^\text{aux}$. The outputs $P_1$ generated by $G^C$ and $P_2$ generated by $G^\text{aux}$ are utilized to compute the probability $P_\text{common}$ that indicates if the sample $x$ comes from common classes. $h$ is the instructive threshold computed by $G^C$ for \textcolor{black}{CMMC} to separate common and unknown samples in the target domain. (b) The illustration of self-paced learning by leveraging cross-domain mixup on different training phase.}
	\label{FIG:model_diagram}
\end{figure*}

\begin{figure}
	\centering
		\includegraphics[width=0.45\textwidth]{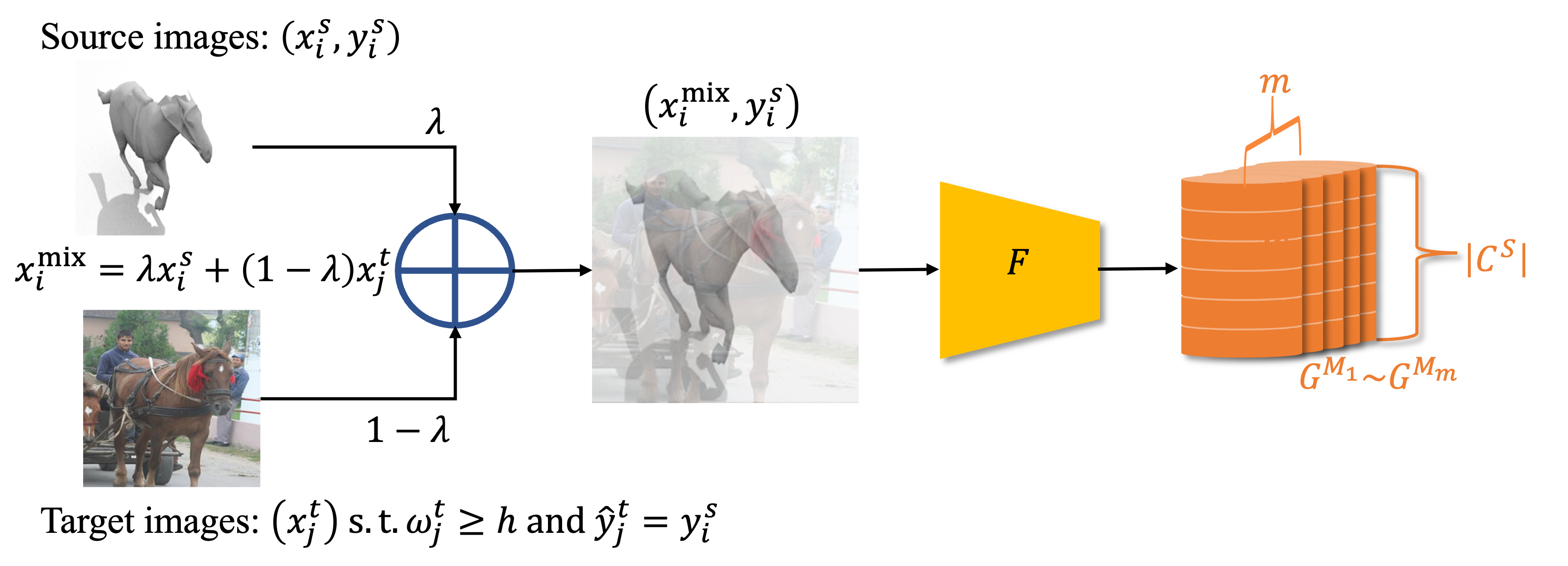}
	\caption{Cross-domain mixup method in \textcolor{black}{CMMC}. $\hat{y}^t_j$ is the pseudo label of the target sample $x^t_j$ predicted by $G^C$. We will mixup $x_j^t$ with same class $x_i^s$ in the pixel level if $\omega_j^t \geq h$. The gradients are not back-propagated to the feature extractor $F$ during \textcolor{black}{CMMC} training phases.}
	\label{FIG:mixup_diagram}
\end{figure}

Fig.\ref{FIG:model_diagram}(a) illustrates the overall pipeline of SPLOS (self-paced learning for open-set). SPLOS consists of \textcolor{black}{dual multi-class classifier (DMC)} module and \textcolor{black}{multi-criteria discriminator with cross-domain mixup (CMMC)} module. We utilize the DMC module to adapt the common classes in the target domain with the source domain and generate an instructive threshold $h$ for CMMC. The CMMC module is used to distinguish common/unknown classes in the target domain. We design a cross-domain mixup method in the CMMC module so the module can learn domain-invariant features in a more continuous latent space. Fig.\ref{FIG:mixup_diagram} illustrates the implementation of our cross-domain mixup method in the CMMC module. 

\subsection{Dual Multi-class Classifier}

To enhance the alignment performance, we propose a novel module called the dual multi-class classifier (DMC), which calculates probabilities indicating whether each sample belongs to the common classes. The target samples belonging to the common classes and source samples should have high probability values. Therefore, we can adapt the common class in the target domain with the source domain precisely. \textcolor{black}{DMC} consists of an adversarial learning classifier $G^C$ and an auxiliary classifier $G^{\text{aux}}$. 

We define a $(\left|C^S\right|+1)$-dimension probability vector $G^C\left(x\right)$ predicted by $G^C$ as follows:
\begin{equation}
\begin{aligned}
G^C\left(x\right)=\left[p_1^{G^C},p_2^{G^C},\ldots,p_{\left|C^S\right|+1}^{G^C}\right].
\end{aligned}
\end{equation}
The probability for the unknown category is indicated by $\left(\left|C^S\right|+1\right)$-th element, and other elements specify the probability of the corresponding class in $C^S$. The samples of common classes in source and target domains tend to have large $\sum_{c=1}^{\left|C^S\right|}p_c^{G^C}$ and small $p_{\left|C^S\right|+1}^{G^C}$. $G^C\left(x\right)$ is calculated by the following formula:
\begin{equation}
\begin{aligned}
G^C\left(x\right)=\frac{\exp\left(l^{G^C}\right)}{\sum_{c=1}^{\left|C^S\right|+1}{\exp\left(l_c^{G^C}\right)}}.
\end{aligned}
\label{eq:calculated_P(G^C)}
\end{equation}
$l^{G^C}$ is the logit vector projected by $G^C$ from the feature $z$, where $z=F(x)$. The loss function of $G^C$ in the source domain is as follows:
\begin{equation}
\begin{aligned}
\mathcal{E}_{G^C}^S=\mathbb{E}_{x^s\sim p}{\left[\mathcal{L}_{CE}\left(x^s,y^s\right) \right]},
\end{aligned}
\label{eq:G^C_loss_on_source_domain}
\end{equation}
where $\mathcal{L}_{CE}$ is a standard cross-entropy loss function.

In addition to $G^C$, we design an auxiliary $\left|C^S\right|$-category classifier $G^\text{aux}$ to compute a probability of a sample belonging to each common class. $G^\text{aux}\left(x\right)$ is a $\left|C^S\right|$-dimension probability vector, which is defined as:
\begin{equation}
\begin{aligned}
G^\text{aux}\left(x\right)=\left[p_1^{G^\text{aux}},p_2^{G^\text{aux}},\ldots,p_{\left|C^S\right|}^{G^\text{aux}}\right].
\end{aligned}
\end{equation}
We calculate $G^\text{aux}\left(x\right)$ with a leaky-softmax function \cite{ETN}, which restrains that $\sum_{c=1}^{\left|C^S\right|}p_c^{G^\text{aux}}$ (i.e., $P_2(x)$ in Fig.\ref{FIG:model_diagram} or Eq.\ref{eq:similarity_source}) is less than 1. If we calculate $G^\text{aux}\left(x\right)$ with standard softmax function, $P_2(x)$ defined in Eq.\ref{eq:similarity_source} will always be equal to 1. It will cause that we cannot combine $G^C$ and $G^\text{aux}$ to compute $P_\text{common}(x)$ since $P_\text{common}(x)$ will always equal $P_1(x)$. $G^\text{aux}\left(x\right)$ is calculated by the following formula:
\begin{equation}
\begin{aligned}
G^\text{aux}\left(x\right)=\frac{\exp\left(l^{G^\text{aux}}\right)}{\left|C^S\right|+\sum_{c=1}^{\left|C^S\right|}{\exp\left(l_c^{G^\text{aux}}\right)}},
\end{aligned}
\end{equation}
 where $l^{G^\text{aux}}$ is the logit vector projected by $G^\text{aux}$ from the feature $z$. The loss function of $G^\text{aux}$ in the source domain is as follows:
\begin{equation}
\begin{split}
\mathcal{E}_{G^{\text{aux}}}^S=\mathbb{E}_{x^s\sim p}{\left[\mathcal{L}_{BCE}\left(x^s,{y'}^s\right) \right]},
\end{split}
\label{eq:Gaux_loss_on_source_domain}
\end{equation}
where ${y'}^s$ is the ground-truth label with one-hot format. $\mathcal{L}_{BCE}$ is a standard binary cross-entropy loss function.

We combine $G^C$ and $G^\text{aux}$ to identify common class samples. $P_\text{common}$ denotes the probability of a sample $x$ belonging to common classes, which can be computed by:
\begin{equation}
\begin{aligned}
P_\text{common}\left(x\right)=P_1(x) \times P_2(x), \\
P_1(x)=\sum_{c=1}^{\left|C^S\right|}p_c^{G^C}, 
P_2(x)=\sum_{c=1}^{\left|C^S\right|}p_c^{G^\text{aux}}.
\label{eq:similarity_source}
\end{aligned}
\end{equation}
We utilize $P_\text{common}$ to optimize $G^C$:

\begin{equation}
\begin{aligned}
\mathcal{E}_{G^C,\text{adv}}^D=\mathbb{E}_{x^t\sim q}\left[-{P_\text{common}\left(x^t\right)}\left(\log{p_{\left|C^S\right|+1}^{G^C}} \right.\right.
\\ \left.\left.+\log{\left(1-p_{\left|C^S\right|+1}^{G^C}\right)}\right)\right] \\ 
+\mathbb{E}_{x^s\sim p}\left[-\left(1-P_\text{common}\left(x^s\right)\right)\left(\log{p_{\left|C^S\right|+1}^{G^C}}\right.\right.
\\ \left.\left.+\log{\left(1-p_{\left|C^S\right|+1}^{G^C}\right)}\right)\right].
\end{aligned}
\label{eq:G^C_domain_adversarial_learning}
\end{equation}
For a target sample $x^t$, only when $P_1(x^t)$ and $P_2(x^t)$ are close to 1, i.e., both $G^C$ and $G^\text{aux}$ agree $x^t$ is from common classes, the model assigns large weight to $x^t$ and aligns $x^t$ with the source domain. The model will not align the target samples with small $P_{\text{common}}(x^t)$ as they only have a minimal contribution to the loss $\mathcal{E}_{G^C,\text{adv}}^D$. For a target sample with a large $P_{\text{common}}(x^t)$, $G^C$ will decrease the first term and the feature extractor $F$ will increase the first term because of the GRL's influence. It can cause $F$ adversarially learns domain-invariant features between target samples from common classes and the source domain. For a source sample $x^s$, its $P_1(x^s)$ and $P_2(x^s)$ should be close to 1 so its weight $\left( 1-P_{\text{common}}\left( x^s \right) \right)$ should be close to 0 but not equal to 0. Our experiment demonstrates that the second term (i.e., the expected loss on the source sample $x^s$) can prevent the model from overconfidently classifying unknown class samples as common classes. We argue that the second term can bring a minor perturbation to source sample features so that unknown samples in the target domain cannot be perfectly aligned with the source domain even though the unknown classes are similar to source classes. We use Nuclear-norm Wasserstein discrepancy \cite{DALN} to train $G^\text{aux}$ to discriminate source and target samples as follows:
\begin{equation}
\begin{aligned}
\mathcal{E}_{G^\text{aux}}^D=\mathbb{E}_{x^s\sim p, x^t\sim q}\left[\left\Vert G^\text{aux}\left(x^t\right) \right\Vert_* - \left\Vert G^\text{aux}\left(x^s\right) \right\Vert_* \right],
\end{aligned}
\label{eq:G^aux_domain_optimization}
\end{equation}
where $\left\Vert\cdot\right\Vert_*$ indicates the Nuclear norm. Notably, as shown in Fig.\ref{FIG:model_diagram}, we do not backpropagate the gradient to the feature extractor when we use Eq.\ref{eq:G^aux_domain_optimization} to optimize the domain discriminability of $G^\text{aux}$. If we backpropagate the gradient to the feature extractor without GRL, it will catastrophically disrupt its ability to generate domain-invariant features. If we backpropagate the gradient to the feature extractor with GRL, the feature extractor will align the whole target domain with the source domain, which leads to negative transfer.

The optimization objectives of the DMC module can be formulated as:
\begin{equation}
\begin{aligned}
\theta_{G^C}^\ast=\arg{\min_{\theta_{G^c}}{\mathcal{E}_{G^C}^S}+\mathcal{E}_{G^C,\text{adv}}^D}, \\
\theta_F^\ast=\arg{\min_{\theta_F}{\mathcal{E}_{G^C}^S}-\mathcal{E}_{G^C,\text{adv}}^D}, \\ \theta_{G^{\text{aux}}}^\ast=\arg{\min_{\theta_{G^{\text{aux}}}}{\mathcal{E}_{G^{\text{aux}}}^S}+\mathcal{E}_{G^{\text{aux}}}^D}.
\end{aligned}
\label{eq:DMC_optimization_objective}
\end{equation}

\subsection{Self-Tunes Threshold}
Before introducing how to compute the threshold, let us consider three potential situations in the target domain: 1) The potential number of unknown samples is much larger than that of common classes. The threshold needs to be high to prevent classifying numerous unknown samples into common classes. 2) The target domain probably contains many samples of common classes and only a few unknown samples. In this case, the threshold tends to be low. Otherwise, a vast number of common class samples will be mislabeled with unknown classes. 3) The potential numbers of common class samples and unknown class samples are comparable. We consider setting the threshold as an intermediate value between the cases of 1) and 2).

Incorporating self-paced learning requires careful consideration of the training process, as it plays a vital role in tuning the threshold. Our goal is for the model to gradually learn the common class features of the target domain, transitioning from easier to more complex examples. By employing a self-tuning threshold that progressively decreases to a suitable value, we can use this threshold as a reference for gradually inputting common class target domain samples with pseudo-labels into the model. This approach enables the model to learn domain-invariant features through self-paced learning, effectively adapting to the target domain while maintaining a focus on common classes. Assume $\left|C^S\right|=2$, and we have two target samples $x_1^t,\ x_2^t$ belonging to class 1 and class 2, both of which are common classes. Since there is a large gap between source and target domains at the initial stage, the threshold should be close to 1. Hence, only a few target samples, which are extremely similar to source samples, are considered as common classes. As the training progresses, the distribution of source and target domains will be aligned gradually. The threshold will be reduced step by step to classify increasingly more target samples as common classes.

Assume we have two training stages which are indicated as the early and late stage. The target domain is more precisely aligned with the source domain in the late stage. Therefore, we assume to get $G^C_\text{early}\left(x^t_1\right)=\left[0.6,\ 0.3,\ 0.1\right]$, $G^C_\text{early}\left(x^t_2\right)=[0.3,\ 0.6,\ 0.1]$, $G^C_\text{late}\left(x^t_1\right)=\left[0.8,\ 0.1,\ 0.1\right]$, and $G^C_\text{late}\left(x^t_2\right)=\left[0.1,\ 0.8,\ 0.1\right]$. The threshold in the early stage should be larger than that in the late stage since the model in the early stage has a higher entropy.

According to the above discussion, we propose a formula of the threshold $h$ produced by $G^C$ in each epoch:
\begin{equation}
\begin{aligned}
h=1-\mathbb{E}_{x_i^t,x_j^t\sim q} \left[\sum_{c=1}^{\left|C^S\right|}\left( \lambda_1\left(G^C\left(x^t_i\right)  + G^C\left(x^t_j\right)\right) \right.\right.
\\ \left.\left. \odot\left(1-\lambda_1\right)\left(G^C\left(x^t_i\right)+G^C\left(x^t_j\right)\right)\right)\vphantom{\sum_{c=1}^{\left|C^S\right|}}\right],
\end{aligned}
\label{eq:compute_threshold}
\end{equation}
 where $i,j$ indicates the index number of target samples that are randomly selected.
 $\lambda_1\in\left[0.5,\ 1\right]$ is a hyperparameter used to control the reduction speed. If $\lambda_1$ closes to 0.5, the threshold drops more rapidly. If $\lambda_1$ closes to 1, the threshold drops more slowly. Experimental results show that our model is not sensitive to hyperparameter $\lambda_1$. $\odot$ means the element product of two probability vectors. The element product can prevent the threshold from dropping too fast and low and becoming unstable. 
 
\begin{figure}
	\centering
		\includegraphics[width=0.35\textwidth]{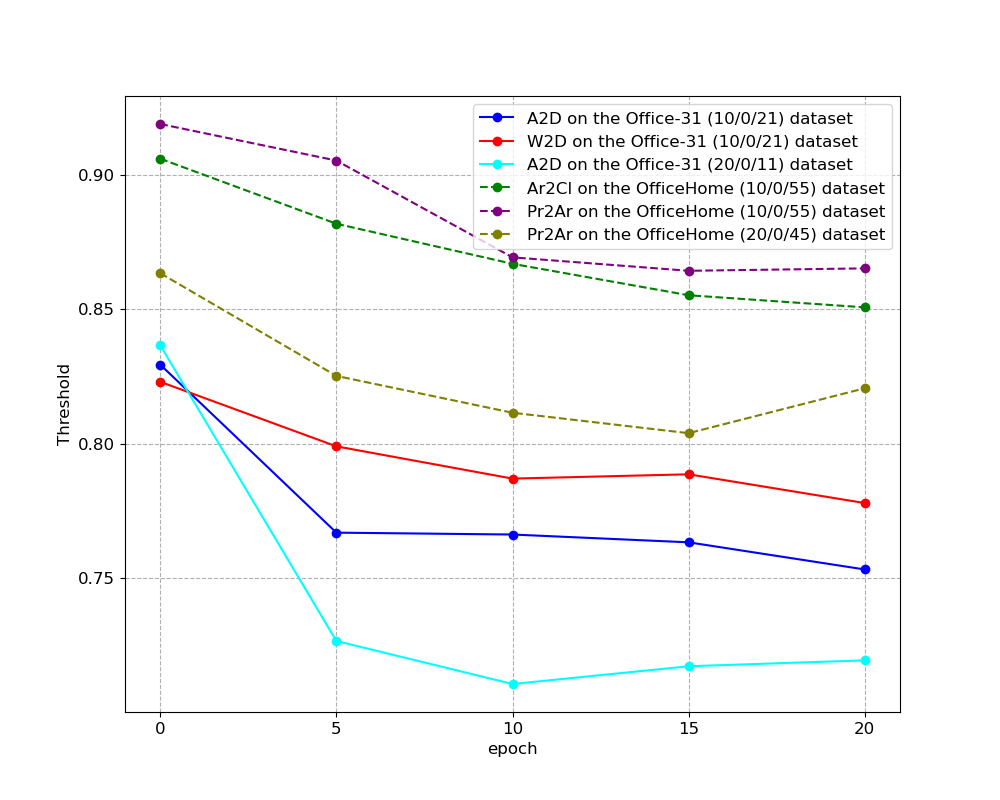}
	\caption{Changes in thresholds on various datasets with different openness.}
	\label{FIG:threshold}
\end{figure}

 Fig.\ref{FIG:threshold} shows how the threshold value changes over time during the training phase on different datasets with varying levels of openness. When the openness $O$ is large,  there is a high probability of choosing two unknown samples $x^t_i, x^t_j$. The first $|C^S|$ terms of the output of $G^C$ will be close to 0 when $x^t_i, x^t_j$ are unknown samples, so we can obtain a large threshold $h$. Conversely, when the openness is small, we will get a smaller threshold value. Furthermore, the first $|C^S|$ terms of $G^C(x^t)$ are small in the initial stage of the training phase since $G^C$ tends to classify most target samples as unknown classes when the target domain is not aligned with the source domain. The threshold $h$ is large in the early stage to prevent mixing unknown class samples with source samples. We will explain the cross-domain mixup method in the next subsection. As the alignment progresses, the threshold will drop gradually until reaching a proper value. 

\subsection{Cross-Domain Mixup with Multiple Criteria}
To implement self-paced learning, we design cross-domain mixup $x^\text{mix}$ samples as input for the cross-domain mixup with multiple criteria (CMMC) module. Fig.\ref{FIG:model_diagram}(b) illustrates the change in $x^\text{mix}$ of a mini-batch. On the one hand, mixing source samples with ground-truth labels and target samples with pseudo-labels minimizes noise caused by pseudo-labels, as we avoid using the entire target samples as input. On the other hand, the source samples provide diverse features during the early training phase when the threshold is high, and only a small quantity of common class samples from the target domain can be utilized. This approach effectively leverages the strengths of both source and target samples, facilitating a smooth progression in the self-paced learning process.

Specifically, we define a $|C^S|$-dimension probability vector $G^{M_k}\left(x\right)$ predicted by each classifier $G^{M_k}$ in CMMC as follows:
\begin{equation}
\begin{aligned}
G^{M_k}\left(x\right)=\left[p_1^{G^{M_k}},p_2^{G^{M_k}},\ldots,p_{|C^S|}^{G^{M_k}}\right].
\end{aligned}
\end{equation}
We calculate $G^{M_k}\left(x\right)$ as the following formula:
\begin{equation}
\begin{aligned}
G^{M_k}(x)=\frac{\exp\left(l^{G^{M_k}}\right)}{\sum_{c=1}^{|C^S|}{\exp\left(l_c^{G^{M_k}}\right)}}.
\end{aligned}
\label{eq:calculated_P(G^{M_k})}
\end{equation}
$l^{G^{M_k}}$ is the logit vector projected by $G^{M_k}$ from the feature $z$, where $z=F(x)$.
As demonstrated in Fig.\ref{FIG:mixup_diagram}, we choose the target samples whose $\omega^t$ is larger than the instructive threshold $h$ given by DMC and mixup them in pixel-level manner with the same class samples in the source domain. We propose the cross-domain mixup method for training each classifier $G^{M_k}$ in CMMC as follows:
\begin{equation}
\begin{aligned}
\mathcal{E}_{G^{M_k}}= \mathbb{E}_{\left(x^s,y^s\right)\sim p,\left(x^t\right)\sim q} \left[\mathcal{L}_{CE}\left(\left(1-\lambda_2\right)x^s +{\lambda_2x}^t,y^s\right)\right] \\ \mathbf{s.t.} \ \ \omega^t\geq h, \hat{y}^t=y^s,
\end{aligned}
\label{eq:loss_with_mixup}
\end{equation}
where $\lambda_2$ is a hyperparameter controlling the ratio of cross-domain mixup and $\hat{y}^t$ is the pseudo label of the target sample $x^t$ predicted by $G^C$.
Furthermore, we propose a more convenient way to automatically adjust $\lambda_2$ for \textcolor{black}{CMMC}. If we have $x_1^t,{\ x}_2^t$ when $\omega_1^t>\omega_2^t>h$, we tend to assign a larger weight $\lambda_2$ to $x_1^t$ since the model has higher confidence in confirming $x_1^t$ comes from the common class. Hence, assigning a higher $\lambda_2$ to $x_1^t$ is less likely to have adverse effects, and the model can further learn the characteristics of the target domain. We retrieve $\lambda_2$ from $\beta$-distribution to implement this intuition. The formal definition is as follows:
\begin{equation}
\begin{aligned}
\lambda_2\sim Beta\left(\omega^t r,hr\right) \ \ \mathbf{s.t.} \ \ \omega^tr>hr>1,
\label{def_extract_lambda_2}
\end{aligned}
\end{equation}
where $r$ is a coefficient to control the probability density function. Thus, the optimization objective of CMMC is as follows:
\begin{equation}
\begin{aligned}
\theta_{G^{M_k}|_{k=1}^m}^\ast=\arg{\min_{G^{M_k}|_{k=1}^m}{\sum_{k=1}^{m}\mathcal{E}_{G^{M_k}}}}.
\end{aligned}
\label{eq:CMMC_optimization_objective}
\end{equation}

Most existing OSDA methods only rely on $confidence$, i.e., the maximum probability mapped by softmax, to determine if a target sample $x^t$ belongs to a specific common class or the unknown class. Although $confidence$ works well when $|C|$ is large, only considering $confidence$ is not a sensible way since it lacks discriminability for the degrees of uncertainty \cite{CMU}. Our experimental results in Fig.\ref{fig:vary_size_H-score} illustrate the performance of \cite{OSBP,DAMC}, which only rely on $confidence$ to distinguish the common and unknown classes, are poor when $|C|$ is small. Therefore, it is necessary to separate common/unknown classes based on various criteria. Inspired by \cite{CMU}, we exploit $entropy, consistency$, and $confidence$ for discriminating common and unknown samples in the target domain.

We utilize probability vectors predicted by $m$ classifiers in CMMC for a target sample $x^t$ to calculate $entropy$ $\omega^\text{ent}$, $consistency$ $\omega^\text{cons}$, and $confidence$ $\omega^\text{conf}$. The formal definitions are as follows:
\begin{equation}
\begin{aligned}
\omega^\text{ent}=\frac{1}{m}\sum_{k=1}^{m}{\sum_{c=1}^{\left|C^S\right|}{-{p_c^{G^{M_k}}}\log{\left({p_c^{G^{M_k}}}\right)}}},
\end{aligned}
\label{eq:compute_entropy}
\end{equation}
\begin{equation}
\begin{aligned}
\omega^\text{cons}=\frac{1}{m\left|C^S\right|}
{\sum_{k=1}^{m}\sum_{c=1}^{\left|C^S\right|}\left(p_c^{G^{M_k}}-\frac{1}{m}\sum_{k=1}^{m}p_c^{G^{M_k}}\right)^2},
\end{aligned}
\label{eq:compute_consistency}
\end{equation}
\begin{equation}
\begin{aligned}
\omega^\text{conf}=\frac{1}{m}\sum_{k=1}^{m}{\max\left(G^{M_k}(x^t)\right)}.
\end{aligned}
\label{eq:compute_confidence}
\end{equation}
$\omega^\text{ent}$ is large for a target sample of target-private classes and small for common classes. $\omega^\text{cons}$ represents the agreement of multiple classifiers $G^{M_k}|_{k=1}^m$, which can compensate for the confidence that usually fails on smooth distribution samples and considers they are uncertain. For smooth distribution, $\omega^\text{cons}$ will be high since classifiers $G^{M_k}|_{k=1}^m$ in CMMC agree with each other. $\omega^\text{conf}$ is high if CMMC more certainly confirms that a target sample is from common classes. Then we integrate $\omega^\text{ent}$, $\omega^\text{cons}$, and $\omega^\text{conf}$ to compute $\omega^t$ for a target sample $x^t$:
\begin{equation}
\begin{aligned}
\omega^t=\frac{\left(1-\omega^\text{ent}\right)+\left(1-\omega^\text{cons}\right)+\omega^\text{conf}}{3}.
\end{aligned}
\label{eq:compute_w^t}
\end{equation}
Smaller $\omega^t$ indicates that $x^t$ is less likely to come from common classes.

Algorithm \ref{alg:SPLOS} outlines the training procedure for SPLOS. To begin, we need to pretrain the CMMC module and feature extractor $F$ only using source samples. The goal of the pretraining phase is to warm up $F$ and the CMMC module, which enables faster convergence and reduces the noise of pseudo-labels during the subsequent training phase. We alternately train DMC and CMMC modules during the training phase. Notably, unlike the pretraining phase, we do not update the parameters of $F$ when we train the CMMC module during the training phase since we wish to keep the outputs of classifiers in CMMC to be diverse in the test phase.

\begin{algorithm}
\caption{The proposed SPLOS}\label{alg:SPLOS}
\begin{algorithmic}[1]
\REQUIRE Labeled source domain $D^S$; Unlabeled target domain $D^T$; Feature extractor $F$; Adversarial learning classifier $G^C$; Auxiliary classifier $G^\text{aux}$; Classifiers $G^{M_1}\sim G^{M_m}$ in the CMMC module.
\ENSURE Learnt networks $F$, $G^C$, $G^\text{aux}$ and $G^{M_1}\sim G^{M_m}$.
\STATE \textbf{Pretraining phase:}
\FOR{$i \gets 1$ to $MaxPreIter$} 
\FOR{$k \gets 1$ to $m$} 
\STATE Sample a batch of source samples $\left\{x^s,y^s\right\} \in D^S$
\STATE Calculate the standard cross-entropy loss of $G^{M_k}$ by $\mathcal{E}_{G^{M_k}}^{\text{pre}}= \mathbb{E}_{\left(x^s,y^s\right)\sim p} \left[\mathcal{L}_{CE}\left(x^s,y^s\right)\right]$
\ENDFOR
\STATE Optimize $F$ and classifiers $G^{M_1}\sim G^{M_m}$ in the CMMC module by $\arg{\min_{G^{M_k}|_{k=1}^m}{\sum_{k=1}^{m}\mathcal{E}^{\text{pre}}_{G^{M_k}}}}$
\ENDFOR
\STATE \textbf{Training phase:}
\FOR{$i \gets 1$ to $MaxEpoches$}
\FOR{$j \gets 1$ to $IterPerEpoch$}
\STATE Sample a batch of source data $\left\{x^s,y^s\right\} \in D^S$ and a batch of target data $\left\{x^t\right\} \in D^T$
\STATE Calculate $\mathcal{E}_{G^C}^S$ by Eq.\ref{eq:G^C_loss_on_source_domain}
\STATE Calculate $\mathcal{E}_{G^{\text{aux}}}^S$ by Eq.\ref{eq:Gaux_loss_on_source_domain}
\STATE Calculate $\mathcal{E}_{G^C,\text{adv}}^D$ by Eq.\ref{eq:G^C_domain_adversarial_learning}
\STATE Calculate $\mathcal{E}_{G^{\text{aux}}}^D$ by Eq.\ref{eq:G^aux_domain_optimization}
\STATE Optimize $F$, $G^C$ and $G^\text{aux}$ by Eq.\ref{eq:DMC_optimization_objective}
\ENDFOR
\STATE Calculate the instructive threshold $h$ by Eq.\ref{eq:compute_threshold}
\FOR{$j \gets 1$ to $IterPerEpoch$}
\FOR{$k \gets 1$ to $m$}
\STATE Sample a batch of source data $\left\{x^s,y^s\right\} \in D^S$ and a batch of target data $\left\{x^t\right\} \in D^T$
\STATE Apply the cross-domain mixup and calculate $\mathcal{E}_{G^{M_k}}$ by Eq.\ref{eq:loss_with_mixup} 
\ENDFOR
\STATE Optimize classifiers $G^{M_1}\sim G^{M_m}$ in the CMMC module by Eq.\ref{eq:CMMC_optimization_objective}
\ENDFOR
\ENDFOR
\RETURN $F$, $G^C$, $G^\text{aux}$ and $G^{M_1}\sim G^{M_m}$
\end{algorithmic}
\end{algorithm}
\begin{figure}
	\centering
		\includegraphics[width=0.45\textwidth]{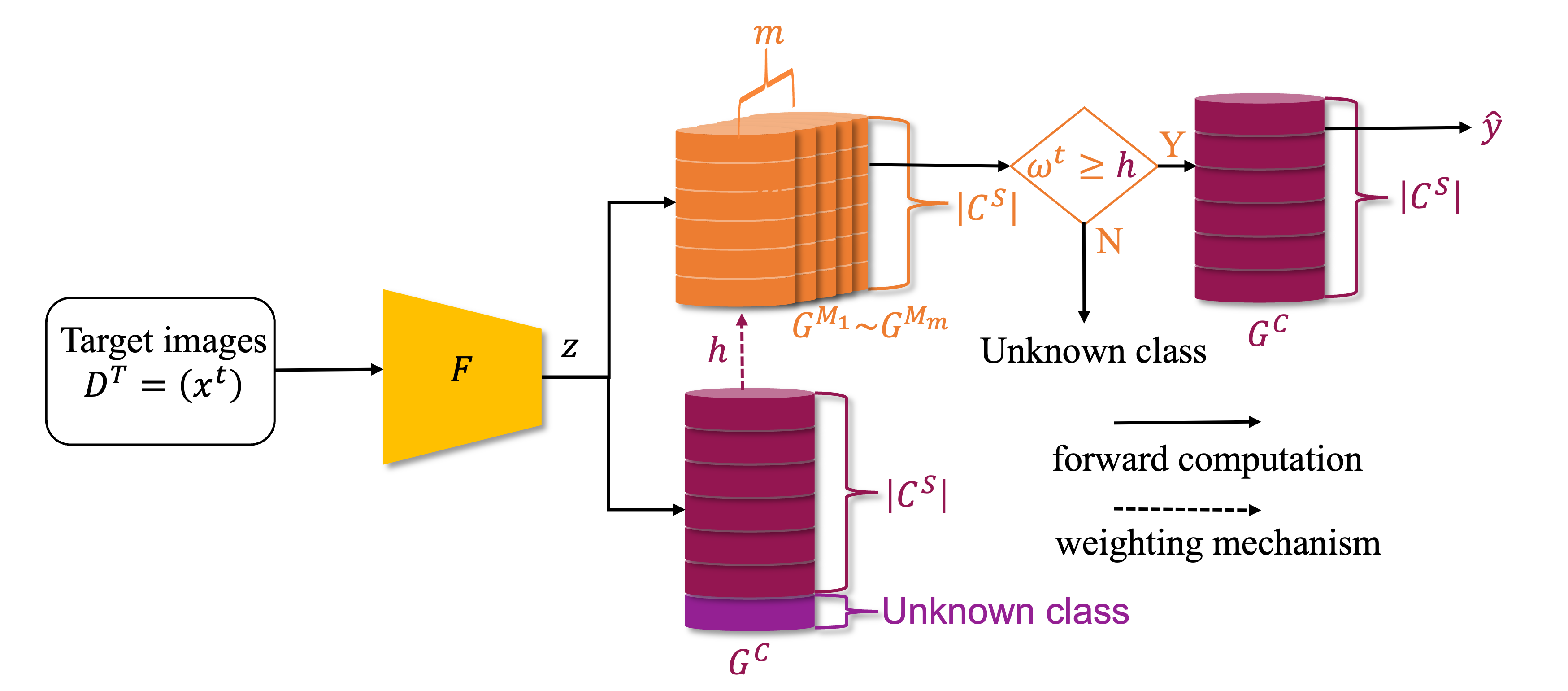}
	\caption{The architecture of SPLOS in the testing phase.} 
	\label{FIG:test_phase_diagram}
\end{figure}
Finally, Fig.\ref{FIG:test_phase_diagram} illustrates the architecture of SPLOS during testing. In the testing phase, we detach the auxiliary classifier $G^\text{aux}$ from SPLOS. We compute the instructive threshold $h$ using Eq.\ref{eq:compute_threshold} instead of setting a threshold manually. SPLOS will classify a target sample $x^t$ as an unknown class if $\omega^t<h$. If the condition is not met, CMMC will further pass $x^t$ to $G^C$, which then classifies $x^t$ into a specific common class.

\begin{table*}[]
\caption{\textbf{OS} (\%) and \textbf{H-score} (\%) on Office-31 (10/0/21) 
 and VisDA-2017 (6/0/6) datasets}
\label{table:office-31_and_visda_results}
\resizebox{1.0\textwidth}{!}{
\begin{tabular}{c|cccccccccccccc|cc}
\hline
               & \multicolumn{14}{c|}{Office-31} & \multicolumn{2}{c}{VisDA-2017}   \\ \cline{2-17} 
Method         & \multicolumn{2}{c}{\textbf{A}$\rightarrow$\textbf{D}} & \multicolumn{2}{c}{\textbf{A}$\rightarrow$\textbf{W}} & \multicolumn{2}{c}{\textbf{D}$\rightarrow$\textbf{A}} & \multicolumn{2}{c}{\textbf{D}$\rightarrow$\textbf{W}} & \multicolumn{2}{c}{\textbf{W}$\rightarrow$\textbf{A}} & \multicolumn{2}{c}{\textbf{W}$\rightarrow$\textbf{D}} & \multicolumn{2}{c|}{Avg}  &  \multirow{2}{*}{OS} &   \multirow{2}{*}{H-score}             \\ \cline{2-15}  & OS                & H-score           & OS                & H-score           & OS                & H-score           & OS                & H-score           & OS                & H-score           & OS                & H-score           & OS             & H-score \\ \hline
DANN \cite{DANN} & 87.83 & 75.78 & 77.13 & 73.16 & 61.79 & 72.25 & 94.48 & 93.91 & 65.09 & 71.07 & 96.23  & 94.89 & 80.43 & 80.18 & 51.31 & 51.41   \\
OSBP \cite{OSBP} & 90.82 & 81.28 & 86.91 & 78.62 & 78.13 & 73.38 & 98.36 & 90.09 & 75.29 & 73.69 & 98.65  & 92.00 & 88.03  & 81.51  & 62.51  & 62.78  \\
DAMC \cite{DAMC} & 90.61 & 81.66 & 86.50 & 84.21 & 76.79 & 79.09 & 97.38 & 89.32 & 75.56 & 79.93 & 97.57 & 84.56 & 87.40 & 83.13 & 40.46 & 47.60 \\
STA \cite{STA} & 88.31 & 39.19 & 89.56 & 49.25 & 78.41 & 58.38 & 94.66 & 58.41 & 74.08 & 56.65 & 93.68 & 46.77 & 86.45 & 51.44 & 60.52 & 51.23\\
PGL \cite{PGL} & 87.91 & 69.44 & 83.19 & 66.45 & 72.42 & 59.22 & 87.84 & 62.94 & 75.04 & 61.06 & 86.26 & 67.84 & 82.11 & 64.49 & \textbf{72.61} & 34.57\\
UADAL \cite{UADAL} & 88.14 & \textbf{87.26} & 84.73 & \textbf{85.59} & 74.99 & 77.11 & 98.33 & 95.03 & 70.23 & 75.86 &99.24 & 95.65 & 85.94 & 86.08 & 59.57 & 61.26 \\
CMU \cite{CMU} & 87.55 & 65.55 & 86.28 & 63.40 & 73.43 & 70.55 & 96.95 & 79.83 & 72.85 & 69.73 & 96.51 & 76.26 & 86.33 & 70.89 & 54.00 & 51.70  \\
DCC \cite{DCC} & 83.14 & 83.65 & 83.99 & 83.35 & \textbf{80.04} & \textbf{81.81} & 96.00 & 92.51 & 75.49 & 79.82 & 98.39 & 90.27 & 86.47 & 84.89 & 59.33 & 59.08  \\
OVANet \cite{OVANet} & 88.19 & 81.77 & 87.62 & 84.30 & 66.02 & 75.51 & 94.23 & 95.62 & 68.56 & 77.60 & 98.94 & 93.83 & 83.93 & 84.77 & 46.13 & 57.53  \\ \hline
SPLOS  & \textbf{91.32} & 84.65 & \textbf{90.20} & 85.05 & 77.43  & 81.29 & \textbf{99.71} & \textbf{98.37} & \textbf{77.15} & \textbf{81.58} & \textbf{99.66} & \textbf{98.07}    & \textbf{89.23} & \textbf{88.17} & 63.59 & \textbf{63.23} \\ \hline
\end{tabular}
}
\end{table*}

\begin{table*}[]
\caption{ \textbf{OS} (\%) and \textbf{H-score} (\%) on the Office-Home (10/0/55) dataset}
\label{table:officehome_results}
\resizebox{1.0\textwidth}{!}{
\begin{tabular}{c|cccccccccccccccccccccccccc}
\hline \multirow{3}{*}{Method}
              & \multicolumn{26}{c}{Office-Home}   \\ \cline{2-27} 
        & \multicolumn{2}{c}{\textbf{Ar}$\rightarrow$\textbf{Cl}} & \multicolumn{2}{c}{\textbf{Ar}$\rightarrow$\textbf{Pr}} & \multicolumn{2}{c}{\textbf{Ar}$\rightarrow$\textbf{Rw}} & \multicolumn{2}{c}{\textbf{Cl}$\rightarrow$\textbf{Ar}} & \multicolumn{2}{c}{\textbf{Cl}$\rightarrow$\textbf{Pr}} & \multicolumn{2}{c}{\textbf{Cl}$\rightarrow$\textbf{Rw}} & \multicolumn{2}{c}{\textbf{Pr}$\rightarrow$\textbf{Ar}} & \multicolumn{2}{c}{\textbf{Pr}$\rightarrow$\textbf{Cl}} & \multicolumn{2}{c}{\textbf{Pr}$\rightarrow$\textbf{Rw}} & \multicolumn{2}{c}{\textbf{Rw}$\rightarrow$\textbf{Ar}} & \multicolumn{2}{c}{\textbf{Rw}$\rightarrow$\textbf{Cl}} & \multicolumn{2}{c}{\textbf{Rw}$\rightarrow$\textbf{Pr}} & \multicolumn{2}{c}{Avg} \\  \cline{2-27}  & OS                & H-score           & OS                & H-score           & OS                & H-score           & OS                & H-score           & OS                & H-score           & OS                & H-score    & OS                & H-score           & OS                & H-score           & OS                & H-score           & OS                & H-score           & OS                & H-score           & OS                & H-score  & OS                & H-score               \\ \hline
DANN \cite{DANN} & 52.72 & 61.94 & 66.23 & 71.86 & 71.39 & 74.79 & 49.79 & 53.48 & 68.47 & 70.46 & 61.06 & 66.50 & 52.05 & 57.55 & 45.83 & 53.51 & 67.81 & 72.39 & 61.87 & 69.31 & 50.86 & 59.55 & 74.98 & 74.66 & 60.25 & 65.50 \\
OSBP \cite{OSBP} & 58.66 & 59.41 & 71.50 & 66.09 & 79.18 & 70.58 & 71.81 & 66.78 & 81.79 & 69.90 & 72.34 & 66.65 & 69.67 & \textbf{66.49} & 53.10 & 54.87 & 78.50 & 69.63 & 73.20 & 69.25 & 60.26 & 56.43 & 78.63 & 67.05 & 70.72 & 65.26 \\
DAMC \cite{DAMC} & 54.04 & 59.40 & 66.26 & 70.10 & 76.48 & 74.20 & 67.65 & \textbf{67.14} & 74.67 & 66.50  & 66.86 & 67.34 & 62.32 & 65.96 & 49.66 & 56.94 & 71.07 & 72.02  & 69.37 & 72.38 & 55.03 & 60.49 & 71.63 & 72.20 & 65.42 & 66.44 \\ 
STA \cite{STA} & 61.47 & 43.50 & 73.64 & 47.96 & 80.30 & 43.94 & 67.19 & 37.62 & 79.90 & 36.68 & 74.94 & 34.78 & 67.61 & 53.41 & \textbf{57.94} & 39.51 & \textbf{78.62} & 49.16 & 75.28 & 49.16 & \textbf{63.22} & 42.66 & 81.82 & 33.84 & 71.83 & 42.69 \\ 
PGL \cite{PGL} & \textbf{64.05} & 53.29 & 75.39 & 58.04 & \textbf{83.23} & 60.77 & \textbf{68.64} & 55.29 & \textbf{82.50} & 59.95 & \textbf{81.69} & 59.89 & \textbf{71.89} & 20.41 & 46.93 & 42.29 & 77.06 & 51.47 & 74.27 & 52.02 & 58.58 & 44.84 & \textbf{82.32} & 51.81 & \textbf{72.21}  & 50.84 \\ 
UADAL \cite{UADAL} & 63.24 & 61.08 & 73.40 & 70.84 & 80.96 & 76.27 & 65.26 & 62.18 & 82.48 & 71.38 & 73.02 & 66.93 & 59.19 & 64.48 & 55.46 & \textbf{60.01} & 73.99 & \textbf{73.73} & \textbf{76.58} & 72.33 & 59.76 & 60.89 & 81.61 & 71.93 & 70.41  & 67.67 \\ 
CMU \cite{CMU} & 45.63 & 55.94 & 59.02 & 68.50 & 67.80 & 74.14 & 46.28 & 55.84 & 57.69 & 65.95 & 58.38 & 66.27 & 42.06 & 52.50 & 39.88 & 49.83 & 60.75 & 68.62 & 60.41 & 68.07 & 46.32 & 55.46 & 68.06 & 73.17 & 54.36 & 62.86\\
DCC \cite{DCC} & 54.45 & 57.03 & \textbf{76.30} & 73.81 & 81.70 & \textbf{79.85} & 56.18 & 35.03 & 68.87 & \textbf{74.62} & 67.52 & 68.93 & 47.18 & 55.74 & 45.87 & 52.60 & 74.01 & 62.08 & 59.73 & 64.63 & 52.31 & 54.62 & 74.00 & 68.87 & 63.18 & 62.32\\
OVANet \cite{OVANet} & 55.03 & 63.20 & 70.27 & 73.03 & 78.64 & 75.11 & 58.05 & 64.15 & 78.91 & 68.63 & 71.40 & 66.74 & 52.12 & 61.02 & 46.82 & 55.66 & 70.97 & 73.27 & 69.34 &\textbf{73.26} & 52.55 & 60.52 & 79.66 & 73.22 &  65.31 & 67.32\\ \hline
SPLOS & 54.48 & \textbf{63.35} & 72.75 & \textbf{74.26} & 78.88 & 78.26 & 62.67 & 66.16 & 74.73 & 72.22 & 68.41 & \textbf{70.14} & 61.93 & 66.30 & 49.19 & 57.30 & 68.27 & 72.63 & 70.03 & 72.68 & 55.38 & \textbf{63.66} & 74.87 & \textbf{77.66} & 65.97 & \textbf{69.55}\\\hline
\end{tabular}
}
\end{table*}

\section{Experiments}

\subsection{Setup}
We choose three benchmarks to evaluate \textcolor{black}{our} model and compare our method with state-of-the-art approaches for open-set domain adaptation on object recognition. We use the label file provided by \cite{jiang2022transferability, tllib} as the label list.

\textbf{Office-31} \cite{Office-31} is a benchmark for domain adaptation, which contains 31 object classes in 3 domains: Amazon (\textbf{A}), DSLR (\textbf{D}) and Webcam (\textbf{W}). We choose the first 10 categories in label order as the common classes and the samples from the remaining 21 classes as private samples of the target domain. The label order in the file is the same as in alphabetical order. 

\textbf{Office-Home} \cite{Office-Home} is a challenging dataset for computer vision domain adaptation, which contains 65 categories in 4 domains: Art (\textbf{Ar}), Clipart (\textbf{Cl}), Product (\textbf{Pr}) and Real-World (\textbf{Rw}). We follow \cite{DAMC} to construct the first 10 classes as the common classes shared by the source and target domains and other categories as the unknown class. We also use label order to sort the classes. The label order in the label file is different from the alphabetical order. Specifically, the first 10 classes are Drill, Exit Sign, Bottle, Glasses, Computer, File Cabinet, Shelf, Toys, Sink, and Laptop. The advantage of following label order is that we can evaluate model performances on the categories even though their first letter is at the end of the Office-Home dataset in alphabetical order.

\textbf{VisDA-2017} \cite{VisDa} focus on adapting synthetic images to real images across 12 classes in 2 domains: synthetic and real. We follow label order to divide 6 categories into common classes for source and target domains and 6 classes to be the private class of the target domain. The label order in the file is the same as in alphabetical order. This setting validates the efficiency of \textcolor{black}{our} model on large-scale domain adaptation tasks.

\textbf{Evaluation Metric.} We employ two metrics to evaluate the performance of different methods: \textbf{OS} and \textbf{H-score} \cite{CMU,ROS}. OS can estimate the average accuracy for $|C|+1$ classes consisting of $|C|$ common classes and the unknown as one class. H-score is the harmonic mean of the accuracy of $OS^{*}$ on common classes $C$ and the accuracy of $Unk$ on the unknown class. The formalization of the H-score is as follows:
\begin{equation}
\begin{aligned}
H_{score}=2\times\frac{OS^{*}\times Unk}{OS^{*}+Unk}
\end{aligned}
\end{equation}

\textbf{Compared State-of-the-Arts.} We have compared SPLOS with: \textbf{1)} Modified closed-set domain adaptation method: DANN proposed in \cite{DANN}, and it was modified by \cite{jiang2022transferability, tllib} so that it can work on OSDA scenarios. \textbf{2)} Open-set domain adaptation methods: OSBP \cite{OSBP}, DAMC \cite{DAMC}, STA \cite{STA}, PGL \cite{PGL} and UADAL \cite{UADAL}. \textbf{3)} Universal domain adaptation (UniDA) methods: CMU \cite{CMU}, DCC \cite{DCC} and OVANet \cite{OVANet}. UniDA methods assume that both source and target domains have private classes, so UniDA methods natively support OSDA tasks.

\textbf{Implementation Details.} We adopt ResNet-50 \cite{ResNet} pre-trained on ImageNet \cite{ImageNet} as the backbone to extract features for fair comparisons. We use Nesterov momentum SGD with the momentum of 0.9 and weight decay of $5\times 10^{-4}$ to optimize our model. We follow \cite{CMU} decaying the learning rate with the factor of $(1+\gamma\times i)^{-\beta}$, where $i$ denotes the current iteration, and we set $\gamma=0.001$ and $\beta=0.75$. $\lambda_1=0.5$ in Eq.\ref{eq:compute_threshold}, $\lambda_2=0.5$ in Eq.\ref{eq:loss_with_mixup}, and $r=30$ in Eq.\ref{def_extract_lambda_2}. We set $m=5$ classifiers in the CMMC module. For Office-31 and VisDA-2017 datasets, the batch size is set to 48. For the Office-Home dataset, the batch size is set to 72. We follow \cite{CMU} to use different data augmentations for \textcolor{black}{CMMC} to enable more diverse classifiers. All experiment is done on $1\times 3090$ GPU. The code is available on GitHub.\footnote{\url{https://github.com/XHomL/SPLOS.git}}

\subsection{Comparison with State-of-the-Arts}

\begin{figure*}[h!]
\centering
      \includegraphics[width=0.3\textwidth]{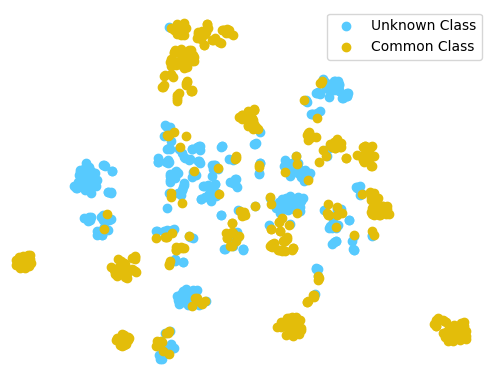}
\hfill
      \includegraphics[width=0.3\textwidth]{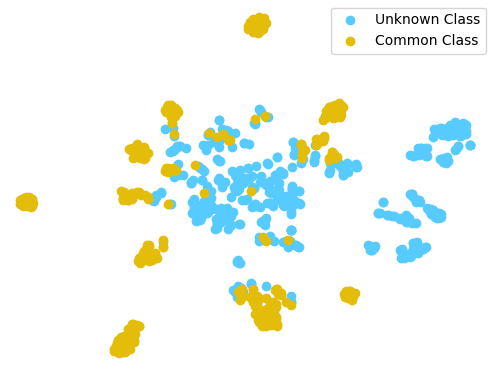}
\hfill
    \includegraphics[width=0.3\textwidth]{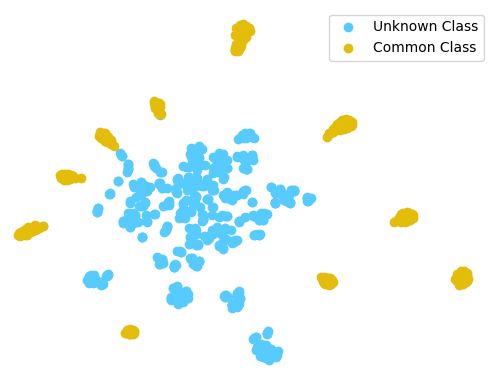}
\hfill
    \subfloat[The top image is annotated by CMU's predictions. The bottom image is annotated by ground truth labels. \label{fig:CMU_pred_and_GT}]{%
      \includegraphics[width=0.3\textwidth]{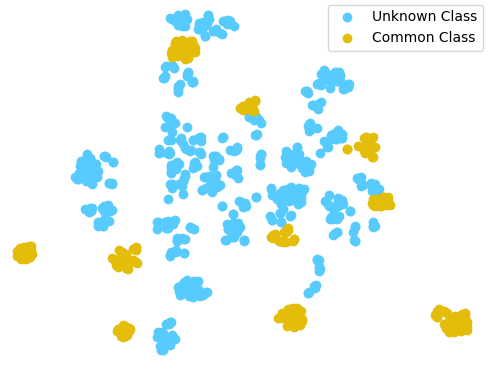}
    }
\hfill
    \subfloat[The top image is annotated by DAMC's predictions. The bottom image is annotated by ground truth labels.\label{fig:DAMC_pred_and_GT}]{%
      \includegraphics[width=0.3\textwidth]{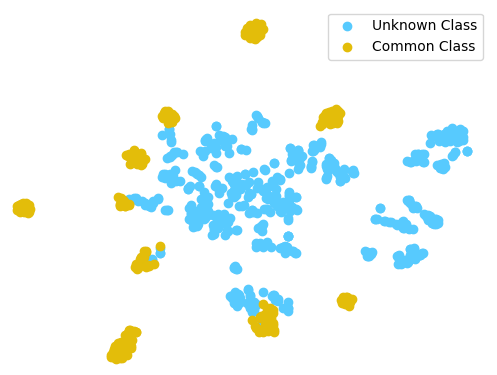}
    }
\hfill
    \subfloat[The top image is annotated by SPLOS's predictions. The bottom image is annotated by ground truth labels.\label{fig:Ours_pred_and_GT}]{%
      \includegraphics[width=0.3\textwidth]{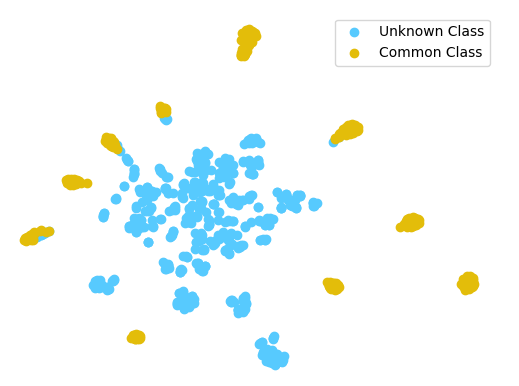}
    }
\caption{An example of feature visualization with t-SNE \cite{t-SNE} for the \textbf{D}$\rightarrow$\textbf{W} task on the Office-31 dataset. We compare our method with CMU \cite{CMU} and DAMC \cite{DAMC} on separating common/unknown target samples. We choose the epoch with the best H-score and visualize target sample predictions. Plots in yellow are common samples, and in blue are unknown samples.}
\label{fig:feature_visualization}
\end{figure*}

\begin{figure*}[h!]
\centering
\subfloat[Amazon to DSLR]{
      \includegraphics[width=0.3\textwidth]{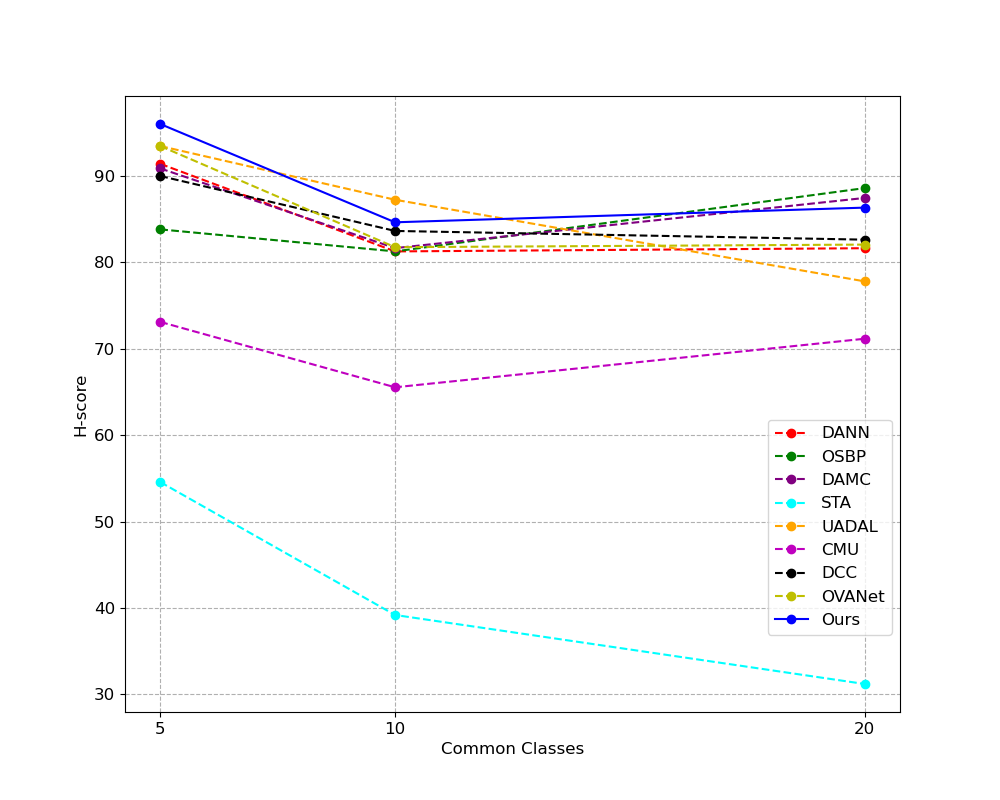}
    }
\hfill
\subfloat[DSLR to Webcam]{
      \includegraphics[width=0.3\textwidth]{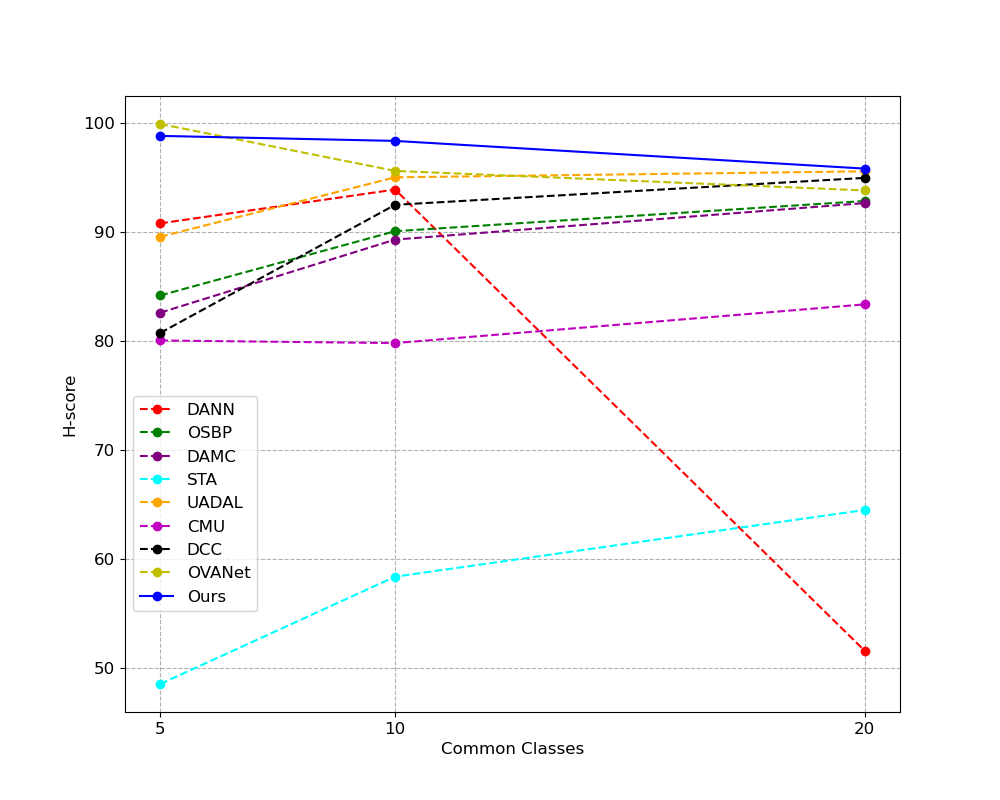}
    }
\hfill
\subfloat[Real World to Clipart]{
      \includegraphics[width=0.3\textwidth]{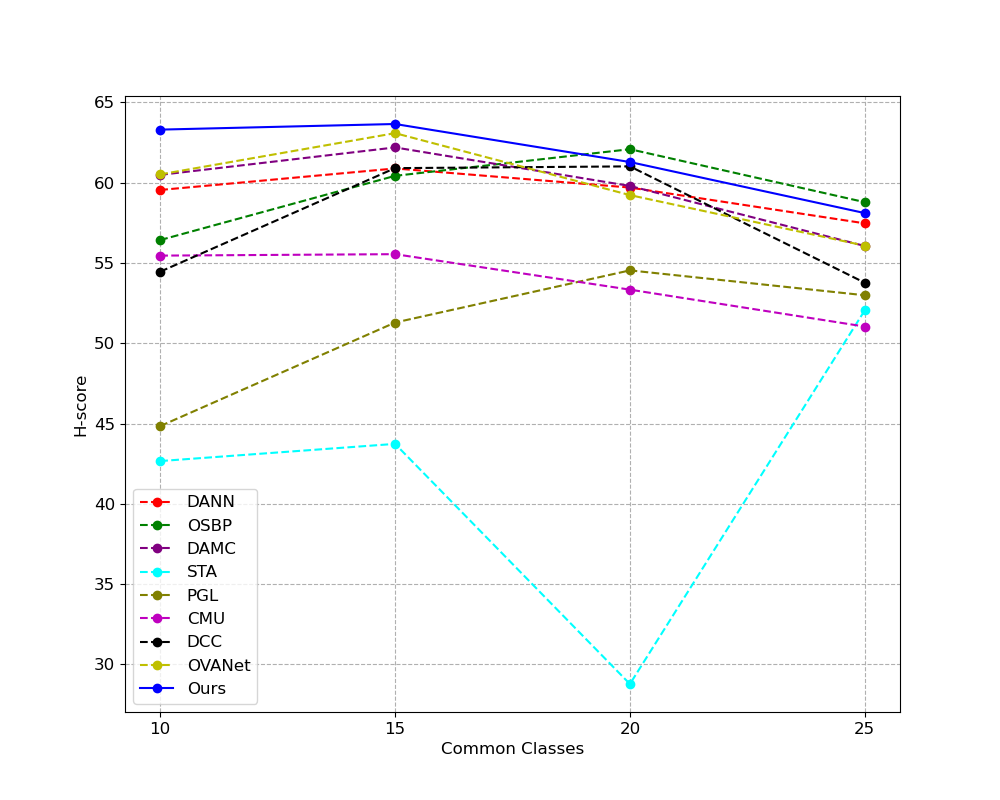}
    }
\caption{H-score in varying sizes of $|C|$. We change $|C|$ on Office-31 and Office-Home.}
\label{fig:vary_size_H-score}
\end{figure*}

The classification results of Office-31 and VisDA-2017 are shown in Table \ref{table:office-31_and_visda_results}, and the results of Office-Home are shown in Table \ref{table:officehome_results}. We compute the H-score on all datasets and OS for Office-31, VisDA-2017, and Office-Home datasets. We choose ResNet-50 \cite{ResNet} as the feature extractor for fair comparisons.

Our proposed method outperforms all the previous methods in terms of OS and H-score on the Office-31 dataset. We can infer that our method works well and exceeds other methods when there is a small domain gap between source and target domains. Moreover, our method outperforms the compared methods on the Office-Home dataset on H-score. The results show that when we deal with OSDA tasks, which are more challenging in large domain gaps and disjoint label space between the source and target domains, SPLOS exceeds the performance of existing OSDA methods by large margins on most tasks of distinguishing common and unknown samples. The results show \textcolor{black}{SPLOS} can adapt the common classes in the target domain to the source domain even though there is a significant gap between source and target domains. We notice that the performance of our model on OS scores still has room for improvement, which will be the research direction of our future work.

Our method consistently performs best and significantly improves OS and H-score on the VisDA-2017 dataset compared to most methods except PGL. PGL has a high OS but a low H-score. It implies that PGL tends to classify a vast number of target-private samples into common classes. In fact, We do not expect this behavior to happen on OSDA tasks. Overall, the results on the VisDA-2017 dataset indicate that SPLOS can learn semantic information from synthetic images and adapt the information to real images shot in diverse scenarios. The results on the VisDA-2017 dataset are more convincing than Office-31 or Office-Home datasets since the dataset contains abundant images. Therefore, random factors on samples only slightly affect the models' performance on OS and H-score. Furthermore, the VisDA-2017 dataset can evaluate the robustness of methods for classifying images with complex scenes.

\begin{figure}
	\centering
		\includegraphics[width=0.35\textwidth]{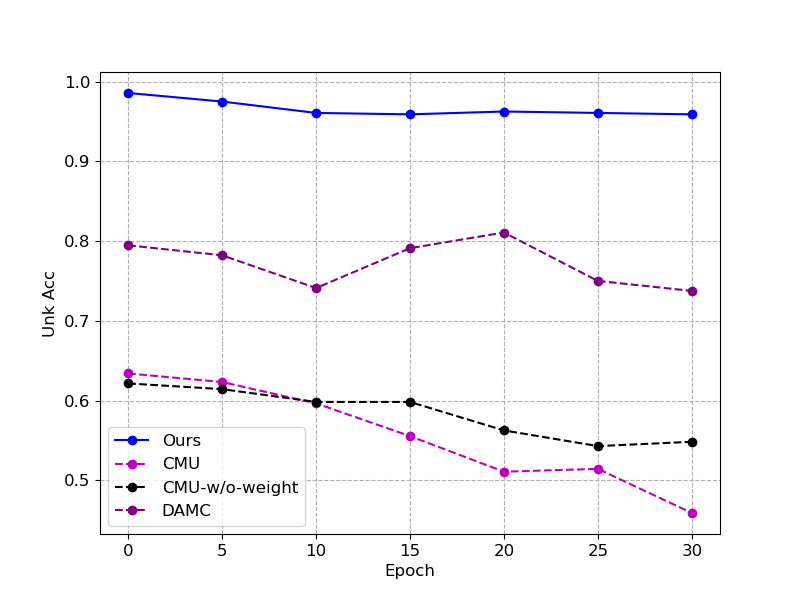}
	\caption{The unknown class accuracy of different methods for the \textbf{D}$\rightarrow$\textbf{W} task on the Office-31 dataset. CMU \cite{CMU} and DAMC \cite{DAMC} are compared. CMU-w/o-weight denotes removing sample weights $\omega^s$ and $\omega^t$ in the domain adversarial discriminator of CMU.}
	\label{FIG:Unk_comparison}
\end{figure}

\subsection{Analysis}

\textbf{Distinguishing common and unknown data.} In order to reduce the marginal distribution divergence between source and target domains, Ganin $et\ al.$ \cite{DANN} construct an adversarial network with domain-hard labels to fool the domain discriminator and then align the whole target domain with the source domain. Nevertheless, those intuitive ideas designed for close-set domain adaptation will induce the models to match the target-private classes with the common classes in OSDA scenarios. Hence, mass unknown class samples will be classified as common classes. Some researchers \cite{CMU,UAN} proposed to compute weights $\omega^s$ for each source sample and $\omega^t$  for each target sample, then deliver to the explicit domain discriminator to distinguish common and unknown classes more accurately. Fig.\ref{fig:feature_visualization} shows an example of the feature visualization with t-SNE \cite{t-SNE} for the \textbf{D}$\rightarrow$\textbf{W} task on the Office-31 dataset. Fig.\ref{FIG:Unk_comparison} is the unknown class accuracy of SPLOS, CMU \cite{CMU} and DAMC \cite{DAMC} in the domain adaptation task \textbf{D}$\rightarrow$\textbf{W} on the Office-31 dataset. The performance of CMU \cite{CMU} in separating common and unknown classes is still poor, as shown in Fig.\ref{fig:CMU_pred_and_GT}. Furthermore, CMU is less stable than the model without assigning weights $\omega^s$ and $\omega^t$ as Fig.\ref{FIG:Unk_comparison} shows. As shown in Fig.~\ref{fig:CMU_pred_and_GT}, \ref{fig:Ours_pred_and_GT} and \ref{FIG:Unk_comparison}, our model performs better in distinguishing common/unknown classes than models with domain-hard labels like CMU \cite{CMU} that induce aligning the whole target domain with the source domain. The domain-hard labels lead to matching the whole target domain with the source domain, and it causes the model to label numerous target-private samples to common classes as shown in Fig.\ref{fig:CMU_pred_and_GT}. The contributions of allocating weights $\omega^s$ and $\omega^t$ are just a drop in the bucket, and the model will become less stable. Compared with DAMC \cite{DAMC}, which only relies on the confidence to distinguish common and unknown classes, we construct the CMMC module to exploit various information. Comparing Fig.\ref{fig:DAMC_pred_and_GT}, \ref{fig:Ours_pred_and_GT} and \ref{FIG:Unk_comparison}, we can conclude that our method performs better and is more stable than DAMC in separating common/unknown classes. The results indicate that our method can precisely identify common class samples from source and target domains and matches them, so it can avoid classifying target-private samples as common classes.

\textbf{Varying Sizes of the Number of Common Classes $|C|$.} The performances of the H-score across various numbers of common classes $|C|$ on Office-31 and Office-Home datasets are visualized in Fig.\ref{fig:vary_size_H-score}. When there are more unknown classes, correctly separating target classes relying only on confidence values is hard. SPLOS overcomes the drawback of OSBP \cite{OSBP} and DAMC \cite{DAMC}, and it works well when the openness $O$ is large. Furthermore, we notice some models, e.g. UADAL, OSBP, DAMC, STA, do not perform well on \textbf{D}$\rightarrow$\textbf{W} OSDA tasks if $|C|$=5. It implies that when two domains are similar and $O$ is large, relying on the confidence is not a reliable way to separate common/unknown samples. The models will recognize unknown samples as common class samples. The performances of SPLOS are also better for most methods when we increase $|C|$. As we mentioned above, the models relying on the confidence can achieve good performances on OSDA tasks if we have a small openness $O$. SPLOS can provide comparable performances when we reduce $O$, which means SPLOS is more general on various $O$ on different OSDA tasks. For most tasks, CMU \cite{CMU} and DANN \cite{DANN} work poorly on different $O$. It indicates that domain-hard labels will negatively impact open-set domain adaptation.

\begin{table*}
\caption{\textbf{OS} ($\%$) and \textbf{H-score} ($\%$) of ablation studies on Office-31 (10/0/21) 
and VisDA-2017 (6/0/6) dataset}
\label{table:office-31_and_visda_ablation_study}
\resizebox{1.0\textwidth}{!}{
\begin{tabular}{c|cccccccccccccc|cc}
\hline
               & \multicolumn{14}{c|}{Office-31} & \multicolumn{2}{c}{VisDA-2017}   \\ \cline{2-17} 
Method         & \multicolumn{2}{c}{\textbf{A}$\rightarrow$\textbf{D}} & \multicolumn{2}{c}{\textbf{A}$\rightarrow$\textbf{W}} & \multicolumn{2}{c}{\textbf{D}$\rightarrow$\textbf{A}} & \multicolumn{2}{c}{\textbf{D}$\rightarrow$\textbf{W}} & \multicolumn{2}{c}{\textbf{W}$\rightarrow$\textbf{A}} & \multicolumn{2}{c}{\textbf{W}$\rightarrow$\textbf{D}} & \multicolumn{2}{c|}{Avg}  &  \multirow{2}{*}{OS} &   \multirow{2}{*}{H-score}             \\ \cline{2-15}  & OS                & H-score           & OS                & H-score           & OS                & H-score           & OS                & H-score           & OS                & H-score           & OS                & H-score           & OS             & H-score \\ \hline
$G^C\text{-w/o-2nd-term}$ & 92.72 & 82.60 & 92.70 & 84.54 & 76.32 & 78.44 & 99.22 & 97.09 & 77.12 & 80.27 & 99.18 & 95.28 & 89.54 & 86.37 & 67.02 & 61.91 \\
SPLOS-w/o-$G^{\text{aux}}$ & 90.17 & 84.43 & 92.58 & 86.37 & 72.60 & 78.75 & 99.63 & 97.90 & 75.21 & 81.67 & 99.71 & 98.38 & 88.32 & 87.92 & 61.41 & 62.99 \\
SPLOS-w/o-CMMC & 92.25 & 86.96 & 85.62 & 85.42 & 64.78 & 74.23 & 91.75 & 94.67 & 65.97 & 75.76 & 99.26 & 98.94 & 83.27 & 86.00 & 51.64 & 59.21\\
SPLOS-w/o-CMMC-$h$ & 82.56 & 81.48 & 89.56 & 80.60 & 75.39 & 78.07 & 98.03 & 94.39 & 73.83 & 78.14 & 99.10 & 94.80 & 86.41 & 84.58 & 62.98 & 60.07 \\
SPLOS-w/o-mixup & 90.88 & 85.38 & 88.60 & 85.46 & 74.48 & 80.93 & 99.69 & 98.27 & 73.22 & 80.44 & 99.66 & 98.07 & 87.76 & 88.09 & 62.95 & 63.20 \\
SPLOS-w-$\beta$ & 91.96 & 85.10 & 90.82 & 84.19 & 76.13 & 81.44 & 99.68 & 98.18 & 75.32 & 81.14 & 99.60 & 97.77 & 88.92 & 87.97 & 63.55 & 62.13 \\
SPLOS  & 91.32 & 84.65 & 90.20 & 85.05 & 77.43  & 81.29 & 99.71 & 98.37 & 77.15 & 81.58 & 99.66 & 98.07  & 89.23 & 88.17 & 63.59 & 63.23   \\ 
\hline
\end{tabular}
}
\end{table*}

\begin{table*}[]
\caption{\textbf{OS} (\%) and \textbf{H-score} (\%) of ablation studies on the Office-Home (10/0/55) dataset}
\label{table:officehome_ablation_study}
\resizebox{1.0\textwidth}{!}{
\begin{tabular}{c|cccccccccccccccccccccccccc}
\hline \multirow{3}{*}{Method}
              & \multicolumn{26}{c}{Office-Home}   \\ \cline{2-27} 
        & \multicolumn{2}{c}{\textbf{Ar}$\rightarrow$\textbf{Cl}} & \multicolumn{2}{c}{\textbf{Ar}$\rightarrow$\textbf{Pr}} & \multicolumn{2}{c}{\textbf{Ar}$\rightarrow$\textbf{Rw}} & \multicolumn{2}{c}{\textbf{Cl}$\rightarrow$\textbf{Ar}} & \multicolumn{2}{c}{\textbf{Cl}$\rightarrow$\textbf{Pr}} & \multicolumn{2}{c}{\textbf{Cl}$\rightarrow$\textbf{Rw}} & \multicolumn{2}{c}{\textbf{Pr}$\rightarrow$\textbf{Ar}} & \multicolumn{2}{c}{\textbf{Pr}$\rightarrow$\textbf{Cl}} & \multicolumn{2}{c}{\textbf{Pr}$\rightarrow$\textbf{Rw}} & \multicolumn{2}{c}{\textbf{Rw}$\rightarrow$\textbf{Ar}} & \multicolumn{2}{c}{\textbf{Rw}$\rightarrow$\textbf{Cl}} & \multicolumn{2}{c}{\textbf{Rw}$\rightarrow$\textbf{Pr}} & \multicolumn{2}{c}{Avg} \\  \cline{2-27}  & OS                & H-score           & OS                & H-score           & OS                & H-score           & OS                & H-score           & OS                & H-score           & OS                & H-score    & OS                & H-score           & OS                & H-score           & OS                & H-score           & OS                & H-score           & OS                & H-score           & OS                & H-score  & OS                & H-score               \\ \hline
$G^C\text{-w/o-2nd-term}$ & 55.98 & 61.95 & 73.29 & 71.88 & 80.15 & 76.31 & 66.00 & 66.84 & 72.33 & 70.86 & 71.01 & 68.52 & 59.53 & 65.98 & 48.91 & 56.52 & 68.30 & 73.05 & 66.91 & 71.63 & 55.55 & 61.53 & 74.69 & 77.40  & 66.05 & 68.54 \\
SPLOS-w/o-$G^{\text{aux}}$ & 54.71 & 62.30 & 72.44 & 73.54 & 79.64 & 77.40 & 64.47 & 66.89 & 73.92 & 72.26 & 69.83 & 70.09 & 60.94 & 65.90 & 47.30 & 56.83 & 67.08 & 73.10 & 67.64 & 72.46 & 55.36 & 62.20 & 73.56 & 77.93 & 65.57 & 69.24 \\
SPLOS-w/o-CMMC & 45.28 & 55.85 & 62.87 & 71.51 & 68.42 & 75.46 & 48.91 & 59.13 & 69.23 & 73.53 & 61.36 & 68.33 & 43.19 & 53.91 & 38.90 & 49.01 & 61.39 & 70.82 & 59.93 & 69.30 & 46.51 & 57.25 & 71.13 & 76.44 & 56.43 & 65.05 \\
SPLOS-w/o-CMMC-$h$ & 56.10 & 59.38 & 73.04 & 70.81 & 79.92 & 75.44 & 69.58 & 64.52 & 78.73 & 64.32 & 73.96 & 64.04 & 64.49 & 64.87 & 52.22 & 54.63 & 75.41 & 68.69 & 74.30 & 69.51 & 57.38 & 57.39 & 78.36 & 70.65 & 69.45 & 65.35\\
SPLOS-w/o-mixup & 51.33 & 60.97 & 69.21 & 75.02 & 77.46 & 78.69 & 61.34 & 66.86 & 71.09 & 72.06 & 71.88 & 70.46 & 55.61 & 64.10 & 45.50 & 54.64 & 68.62 & 74.14 & 66.29 & 73.00 & 55.46 & 62.97 & 74.58 & 78.29 & 64.03 & 69.27 \\
SPLOS-w-$\beta$ & 55.05 & 63.39 & 71.14 & 74.01 & 79.58 & 78.15 & 63.92 & 66.69 & 72.87 & 72.08 & 69.25 & 69.56 & 59.93 & 65.07 & 47.49 & 57.11 & 65.97 & 72.48 & 69.54 & 72.94 & 57.23 & 63.02 & 72.10 & 77.24 & 65.34 & 69.32 \\
SPLOS & 54.48 & 63.35 & 72.75 & 74.26 & 78.88 & 78.26 & 62.67 & 66.16 & 74.73 & 72.22 & 68.41 & 70.14 & 61.93 & 66.30 & 49.19 & 57.30 & 68.27 & 72.63 & 70.03 & 72.68 & 55.38 & 63.66 & 74.87 & 77.66 & 65.97 & 69.55\\
\hline
\end{tabular}
}
\end{table*}
\textbf{Ablation Studies.} We evaluate the effectiveness of the proposed components in \textcolor{black}{SPLOS} on Office-31, Office-Home and VisDA-2017 datasets separately. 1) $G^C\text{-w/o-2nd-term}$ means that we remove the second term in Eq.\ref{eq:G^C_domain_adversarial_learning}. 2) SPLOS-w/o-$G^\text{aux}$ represents that we remove $G^\text{aux}$ in the DMC module and we rely only $G^C$ to compute the weight in Eq.\ref{eq:G^C_domain_adversarial_learning} to align target and source domains. In this case, $P_{\text{common}}(x)=P_1(x)$ in Eq.\ref{eq:G^C_domain_adversarial_learning} instead of $P_{\text{common}}(x)=P_1(x)\times P_2(x)$. 3) SPLOS-w/o-CMMC denotes that we remove the CMMC module and use the output of $G^C$ to predict the target samples' classes. If $\forall p_c^{G^C}\geq h, c\in \left[1,|C^S|\right]$, we classify the target sample as the common class $c$. Otherwise, we predict the target sample as unknown. 4) SPLOS-w/o-CMMC-$h$ is similar to the 3) case, but we use the maximum confidence $\max \left(p_c^{G^C}\right), c\in \left[1,|C^S|+1\right]$ rather than compare the confidence with the threshold to predict the target sample's class $c$. 5) SPLOS-w/o-mixup represents a variation of our method in which the cross-domain mixup is not used in the CMMC module, meaning that self-paced learning is not implemented. This variant serves as a baseline for comparison, allowing us to evaluate the impact of incorporating self-paced learning in our proposed approach. 6) SPLOS-w-$\beta$ represents that we let the model automatically retrieve $\lambda_2$ in Eq.\ref{eq:calculated_P(G^{M_k})} from Eq.\ref{def_extract_lambda_2}.

The results presented in Tables \ref{table:office-31_and_visda_ablation_study} and \ref{table:officehome_ablation_study} demonstrate the effectiveness of the second term in Eq.\ref{eq:G^C_domain_adversarial_learning} for helping the model distinguish between common and unknown samples. This term prevents the model from erroneously classifying unknown samples as common classes. The auxiliary classifier $G^\text{aux}$ contributes to calculating a more suitable weight $P_\text{common}$, enabling more precise alignment between the two domains. Comparing the performances of SPLOS-w/o-CMMC and SPLOS-w/o-CMMC-$h$, it is evident that combining entropy, consistency, and confidence offers a more reliable approach for distinguishing common and unknown classes than relying solely on confidence. The results for SPLOS-w/o-mixup suggest that the self-paced learning by leveraging the cross-domain mixup method assists SPLOS in learning domain-invariant features within a more continuous feature space, particularly when a significant gap exists between the source and target domains. The comparable performance between \textcolor{black}{SPLOS-w-$\beta$} and SPLOS indicates that SPLOS can effectively extract the suitable mixup parameter $\lambda_2$ from the $\beta$-distribution by itself. This finding further validates the robustness and adaptability of our proposed approach in handling open-set domain adaptation tasks.

\begin{figure}
	\centering
		\includegraphics[width=0.45\textwidth]{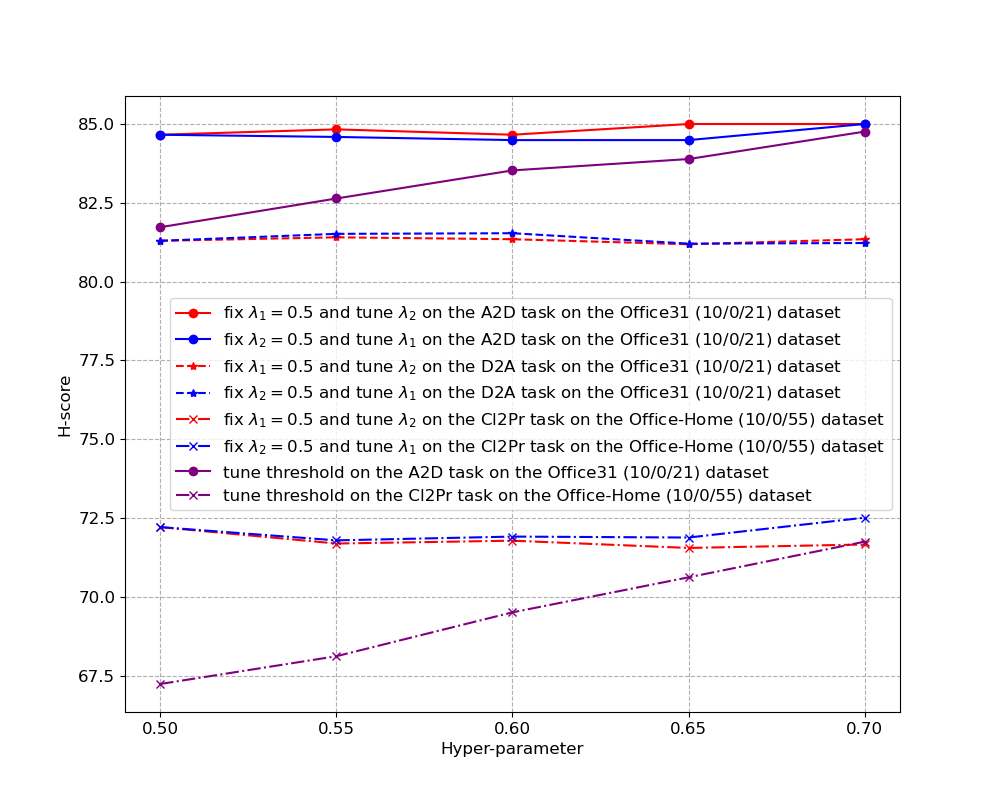}
	\caption{Sensitivity of hyperparameters $\lambda_1$ and $\lambda_2$. $\lambda_1$ is for controlling the rate of descent of threshold $h$ in Eq.\ref{eq:compute_threshold}, and $\lambda_2$ is for the mixup ratio in Eq.\ref{eq:loss_with_mixup}.}
	\label{FIG:vary_lambda}
\end{figure}
\begin{table}[]
\caption{\textbf{OS} (\%) and \textbf{H-score} (\%) with various $m$ and $\beta$ on Office-31 (10/0/21), Office-Home (10/0/55) and VisDA-2017 (6/0/6) datasets}
\resizebox{0.48\textwidth}{!}{
\begin{tabular}{c|cccccc}
\hline
\multicolumn{1}{c|}{\multirow{2}{*}{Hyperparameters}} & \multicolumn{2}{c}{Office-31} & \multicolumn{2}{c}{Office-Home} & \multicolumn{2}{c}{Visda-2017} \\ \cline{2-7} 
\multicolumn{1}{c|}{}                                  & OS          & H-score         & OS           & H-score          & OS          & H-score          \\ \hline
$m$=3 & 88.55 & 87.99 & 65.49 & 69.47 &  63.21 &  63.38 \\
$m$=4 & 88.54  & 87.72 & 66.37 & 69.23 & 63.15 & 63.72  \\
$m$=5 & 89.23 &  88.17 & 65.97 & 69.55 & 63.59 & 63.23 \\ \hline
$\beta$=10 & 89.10 & 88.03 & 65.97 & 69.23 & 61.22 & 63.38 \\
$\beta$=20 & 89.05 & 87.89 & 65.97 & 69.23 & 61.25 & 63.33  \\
$\beta$=30 & 88.92 & 87.97  & 65.34 & 69.32 & 63.55 & 62.13 \\
\hline
\end{tabular}
}
\label{various_m_beta}
\end{table}
\textbf{Robustness of Hyperparameters.} We tune the hyperparameters $\lambda_1$ in Eq.\ref{eq:compute_threshold} and $\lambda_2$ in Eq.\ref{eq:loss_with_mixup} on \textbf{A$\rightarrow$D} and \textbf{D$\rightarrow$A} tasks of the Office-31 dataset and the \textbf{Cl$\rightarrow$Pr} task of the Office-Home dataset. The results in Fig.\ref{FIG:vary_lambda} indicate that our method is not sensitive to changing hyperparameters $\lambda_1$ and $\lambda_2$. When compared to tuning the threshold, which can significantly impact the model's performance, our model demonstrates stable performance across varying $\lambda_1$ and $\lambda_2$. Hence, our method is free from empirically tuning the optimal threshold. We also explored various numbers of classifiers, represented by $m$, in the CMMC module and $\beta$ in Eq.\ref{def_extract_lambda_2}. The results in Table \ref{various_m_beta} convey that SPLOS maintains steady performance with different $m$ and $\beta$.

\section{Conclusion}
In this paper, we propose a novel Self-Paced Learning for Open-Set Domain Adaptation (SPLOS) framework. SPLOS comprises two key modules: the Dual Multi-Class Classifier (DMC) and the Cross-Domain Mixup with Multiple Criteria (CMMC). The DMC module aligns common classes in the target domain with the source domain, instead of aligning the entire target domain, while utilizing unlabeled target samples to compute a suitable threshold for CMMC. This approach enables the implementation of self-paced learning through cross-domain mixup in the CMMC module. CMMC learns common class features of the target domain, progressing from easier to more complex examples. Our approach self-tunes a suitable threshold, eliminating the need for empirical tuning during testing. Experimental results on three OSDA benchmarks demonstrate that the SPLOS model outperforms state-of-the-art methods, particularly when the openness $O$ is large. Moreover, SPLOS exhibits robustness to hyperparameters and delivers stable performance. In future work, we aim to further improve the OS score of SPLOS on the Office-Home dataset. We also plan to extend our work to tackle universal domain adaptation tasks.


%
\bibliographystyle{IEEEtran}
\bibliography{main}












\vfill

\end{document}